\documentclass[conference]{IEEEtran}
\IEEEoverridecommandlockouts
\usepackage{cite}
\usepackage{amsmath,amssymb,amsfonts}
\DeclareMathOperator*{\argmin}{arg\,min}
\usepackage{algorithmic}
\usepackage{graphicx}
\usepackage{textcomp}
\usepackage{xcolor}
\usepackage[ruled,vlined,linesnumbered]{algorithm2e}
\SetKwComment{Comment}{/* }{ */}
\usepackage{booktabs, multirow}

\usepackage{hyperref}
\usepackage{amssymb}
\usepackage{pifont}
\newcommand{\cmark}{\ding{51}}
\newcommand{\xmark}{\ding{55}}
\usepackage{caption}
\usepackage{tabu}

\def\BibTeX{{\rm B\kern-.05em{\sc i\kern-.025em b}\kern-.08em
    T\kern-.1667em\lower.7ex\hbox{E}\kern-.125emX}}
\begin{document}

\title{CaT: Balanced Continual Graph Learning with Graph Condensation\\

}
\author{\IEEEauthorblockN{Yilun Liu, Ruihong Qiu, and Zi Huang}
\IEEEauthorblockA{\textit{School of Electrical Engineering and Computer Science, The University of Queensland,}\\
\{yilun.liu, r.qiu, helen.huang\}@uq.edu.au}
}

\maketitle

\begin{abstract}
Continual graph learning (CGL) is purposed to continuously update a graph model with graph data being fed in a streaming manner. Since the model easily forgets previously learned knowledge when training with new-coming data, the catastrophic forgetting problem has been the major focus in CGL. Recent replay-based methods intend to solve this problem by updating the model using both (1) the entire new-coming data and (2) a sampling-based memory bank that stores replayed graphs to approximate the distribution of historical data. After updating the model, a new replayed graph sampled from the incoming graph will be added to the existing memory bank. Despite these methods are intuitive and effective for the CGL, two issues are identified in this paper. Firstly, most sampling-based methods struggle to fully capture the historical distribution when the storage budget is tight. Secondly, a significant data imbalance exists in terms of the scales of the complex new-coming graph data and the lightweight memory bank, resulting in unbalanced training. To solve these issues, a \textit{Condense and Train (CaT)} framework is proposed in this paper. Prior to each model update, the new-coming graph is condensed to a small yet informative synthesised replayed graph, which is then stored in a \textit{Condensed Graph Memory} with historical replay graphs. In the continual learning phase, a \textit{Training in Memory} scheme is used to update the model directly with the \textit{Condensed Graph Memory} rather than the whole new-coming graph, which alleviates the data imbalance problem. Extensive experiments conducted on four benchmark datasets successfully demonstrate superior performances of the proposed CaT framework in terms of effectiveness and efficiency. The code has been released on \url{https://github.com/superallen13/CaT-CGL}.
\end{abstract}


\section{Introduction}
\label{sec:intro}

\begin{figure}[!t]
\centering
\begin{tabular}{cc}
  \multicolumn{2}{c}{\includegraphics[width=85mm]{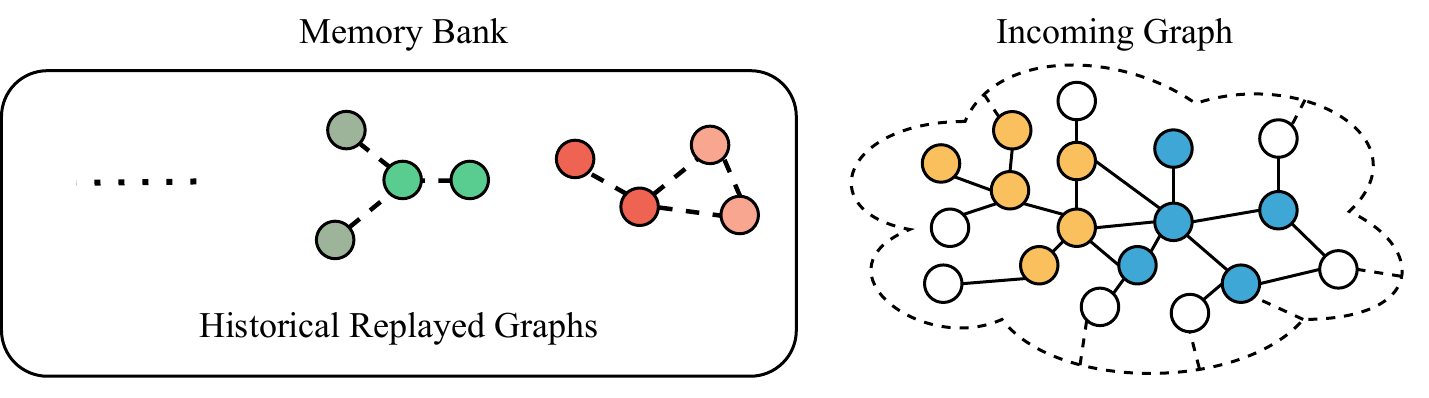}}\\
  \multicolumn{2}{c}{(a) Data imbalance in replay-based CGL methods.}\\
  \includegraphics[width=41mm]{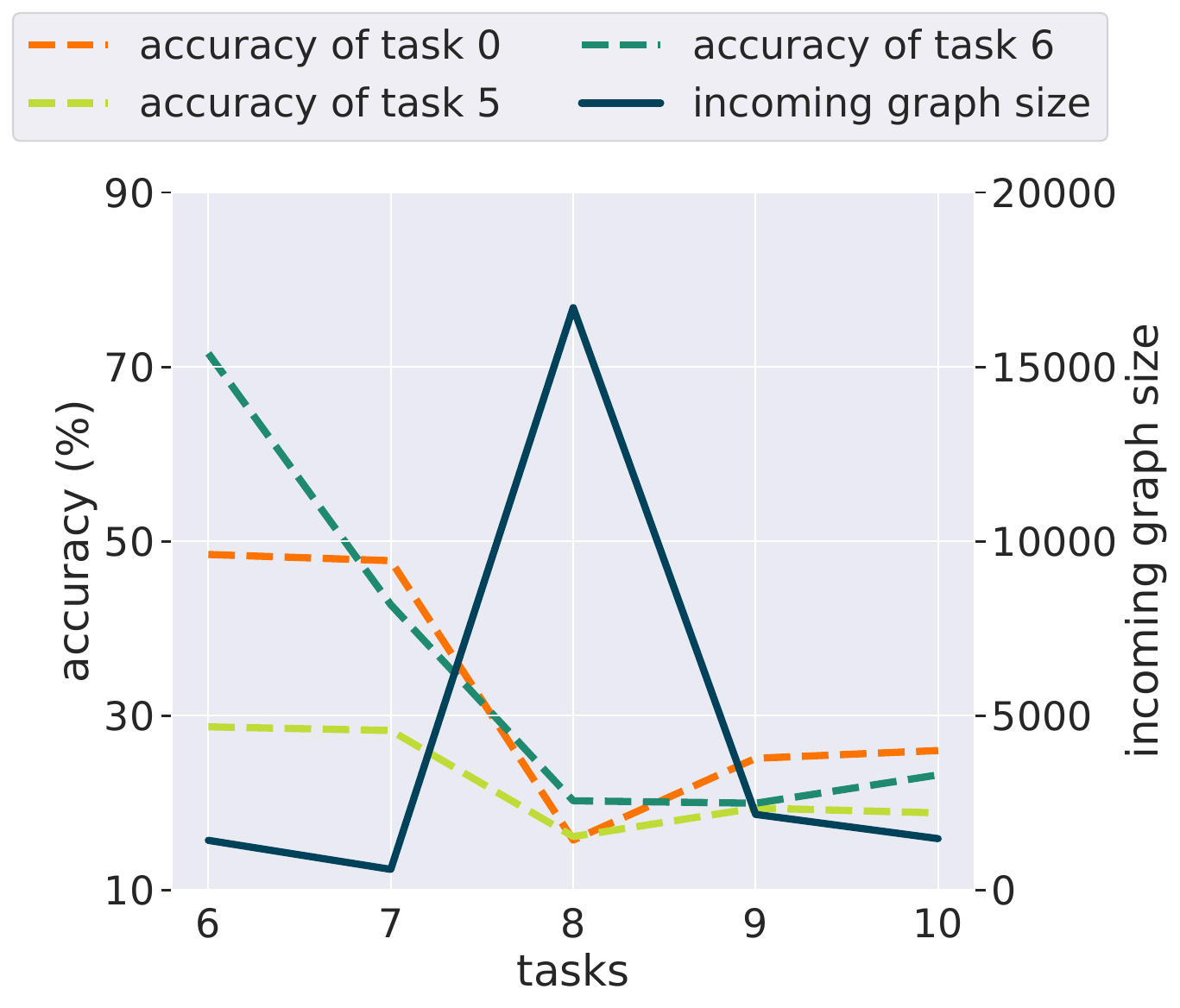} & \includegraphics[width=41mm]{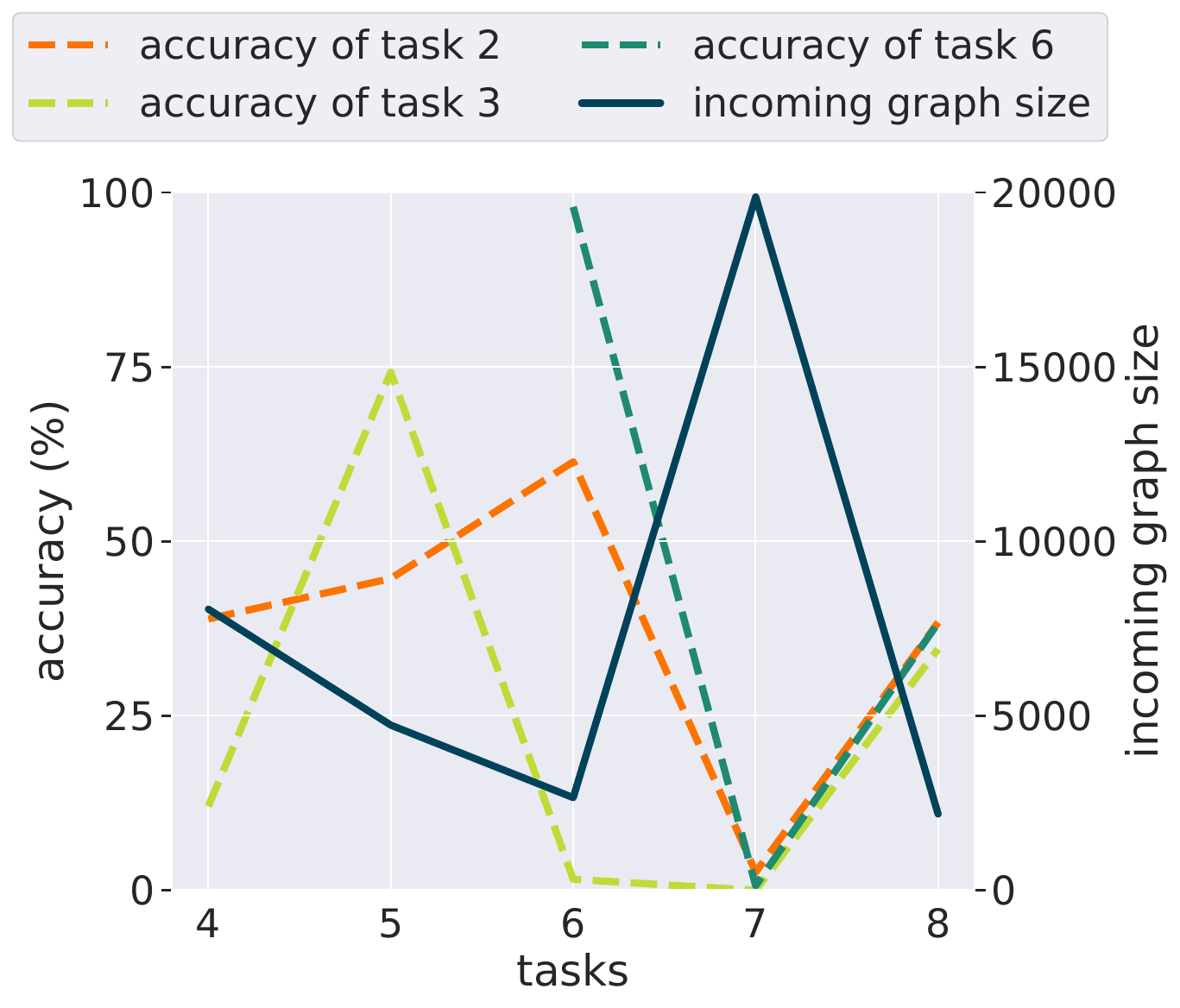} \\
  (b) Imbalance in Arxiv & (c) Imbalance in Reddit
\end{tabular}
\caption{Imbalanced learning problem in the replay-based CGL methods. (a) In the update phase of replay-based methods, both the replayed graphs in the memory bank and the incoming graph are used for training. Generally, the replayed graphs are significantly smaller than the incoming graph. In (b) and (c), the prediction accuracy of the CGL model on Arxiv and Reddit datasets is shown. When the size of the incoming graph is much larger than those of replayed graphs, such as Task 8 in Arxiv and Task 7 in Reddit, the model performance over data drawn from previous tasks will drastically decrease. This is due to the data imbalance issue in the continual learning.}
\label{fig:1}
\end{figure}

Compared to traditional graph representation learning that treats graphs as static data and trains a model with the data as a whole, continual graph learning (CGL) deals with a more practical scenario where the graph data is emerged continually and is fed into the model in a streaming manner~\cite{cgl-recsys1, cgl-kg, gag}. 

For the CGL problem, the most significant challenge is to address a catastrophic forgetting problem that the model easily forgets the knowledge learned from the historical graph data while overemphasising the incoming data~\cite{ergnn,twp,ssm}. Due to hardware limitations in storage and computation, the catastrophic forgetting problem happens to CGL models when it is impractical to access the entire historical graph during training. If conventional Graph Neural Networks (GNNs)~\cite{gcn,gat,sgc} are directly employed to continually learn from incoming graphs, the model performance for data from the historical distribution tends to deteriorate due to the distribution shift between the historical and the current graphs. Recently, a few attempts have been made to tackle this catastrophic forgetting problem, leveraging regularisation penalty~\cite{twp},  architecture redesign~\cite{hpn}, and replayed graph~\cite{ergnn, ssm}. As the champion of the three methods, replay-based CGL models store replayed graphs in a memory bank by sampling methods to maintain the historical distribution to tackle the catastrophic forgetting problem, resulting in improved performance and plasticity. For example, ER-GNN~\cite{twp} stores informative nodes from historical graphs in the memory bank, and SSM~\cite{ssm} sparsifies the incoming graphs as replayed graphs.

Although replay-based methods are intuitive and effective, two major issues are observed during our study of these approaches. Firstly, to achieve a competitive performance, existing replay-based CGL methods usually require large spaces to store the historical information as much as possible. When the storage budget is limited, these memory banks would hardly present the complete picture of the distribution of historical data. Secondly, it is difficult to balance the model update training over incoming and replayed graphs since incoming graphs are generally much larger than replayed graphs in scale. Figure (Fig.)~\ref{fig:1}(a) demonstrates the situation of imbalanced learning where the incoming graph is significantly larger than replayed graphs. Fig.~\ref{fig:1}(b) and Fig.~\ref{fig:1}(c) show that the model performance on the historical data drops when the incoming graph is significantly larger than replayed graphs.

In light of the discussion above, it is motivated to design a novel framework that can simultaneously improve the effectiveness and efficiency of the memory bank and balance the continual training for replay-based GCL methods. To generate small but informative replayed graphs, recent graph condensation techniques~\cite{gc-multi, gc-one, gc-dist} have demonstrated great potential, which can condense a graph into a smaller synthetic graph using differentiable methods. Compared to the sampling-based replayed graphs, the graph condensation has the merits of generating smaller and learnable replayed graphs without compromising the performance. 
Regarding the imbalanced training, it is difficult to maintain a balance when replayed graphs and the entire incoming graph are directly combined to train the model due to the nature of the imbalanced data scale.
Instead, if the synthetic replayed graph derived from condensation methods is able to support the model training without sacrificing the performance, it is possible to bypass the usage of the entire incoming graph for a model update but rather purely rely on the synthetic graphs. To fulfil these two objectives, we propose a replay-based \textbf{\textit{\underline{C}ondense \underline{a}nd \underline{T}rain (CaT)}} framework for CGL. In the continual learning process, it maintains a small yet effective \textbf{\textit{\underline{C}ondensed \underline{G}raph \underline{M}emory (CGM)}} that expands with the synthetic replayed graph condensed from the incoming graph before the model update. In the model update phase, the training strategy used for the proposed framework is \textbf{\textit{\underline{T}raining \underline{i}n \underline{M}emory (TiM)}} where the model only updates with the memory bank. The TiM ensures that the condensed synthetic graph has a similar size to replayed graphs, alleviating the imbalance issue. The contributions are as follows:
\begin{itemize}
    \item For CGL problem, a novel framework CaT is proposed with a CGM module to reduce the size of replayed graphs and a TiM scheme to balance the continual training.
    \item CGM is derived from performing a graph condensation for the large incoming graphs using distribution matching.
    \item TiM is developed to balance the training using the large incoming graph and the small memory bank.
    \item Extensive experiments conducted on four benchmark datasets verify the state-of-the-art performance of CaT.
\end{itemize}

\section{Related Work}
\label{sec:liter}

\subsection{Graph Neural Networks}
Graph Neural Networks (GNNs) are effective tools for graph-based tasks~\cite{gin,ggnn}. GCN~\cite{gcn} employs Laplacian normalisation for message propagation. GAT~\cite{gat} is proposed to use the attention mechanism~\cite{att} for message passing. SGC~\cite{sgc} simplifies GCN by removing the non-linear activation layer. GraphSAGE~\cite{sage} uses node sampling to deal with large-scale graph representation learning.

\subsection{Graph Continual Learning}
Graph continual learning is a task for handling streaming graph data. Continual learning has been studied in computer vision~\cite{ewc,mas,gem,lwf}. In graph area, CGL methods can be categorised into three branches: regularisation~\cite{twp}, replay-~\cite{ergnn, ssm}, and architecture-based~\cite{hpn} methods. TWP~\cite{twp} preserves the topological information of historical graphs by adding regularisation. HPNs~\cite{hpn} redesigns the architecture to 3-layer prototypes for representation learning. ER-GNN~\cite{ergnn} integrates memory-replay by storing representative nodes. SSM~\cite{ssm, zhang2023sufficient} stores the sparisified subgraphs in the memory bank to preserve the structural information. Two recent benchmarks~\cite{cglb, begin} have been developed for CGL.

\subsection{Graph Condensation}
Dataset condensation generates a small and synthetic dataset to replace the original dataset and to train a model with similar performance. Dataset condensation has been applied in computer vision~\cite{dc,dcgm,dsa,cafe,dm,glad}. Recently, gradient matching has been applied to graph condensation, such as GCond~\cite{gc-multi}, DosCond~\cite{gc-one} and MCond~\cite{mcond}. DM~\cite{dm} aims to learn synthetic samples which have similar distribution with the original dataset to mimic sampling methods~\cite{herding, coreset1, coreset2}. GCDM~\cite{gc-dist} uses the distribution matching for graph condensation.

Recent attempts on computer vision have directly applied dataset condensation to continual learning~\cite{cldc1, cldc2, cldc3, cldc4, cldc5}, although these methods follow the typical training scheme, which will fall into imbalanced training.

\section{Preliminary}
\label{sec:pre}

In the following, a bold lowercase letter denotes a vector, a bold uppercase letter denotes a matrix, a general letter denotes a scalar, and a scripted uppercase letter denotes a set.

\subsection{Graph}

For a node classification problem, a graph is denoted as $\mathcal{G}=\{\boldsymbol{A}, \boldsymbol{X}, \boldsymbol{Y}\}$, where $\boldsymbol{X} \in \mathbb{R}^{n \times d}$ is the $d$-dimensional feature matrix for $n$ nodes, and the adjacency matrix $\boldsymbol{A} \in \mathbb{R}^{n \times n}$ denotes the graph structure. In this paper, the graph is undirected and unweighted. $\boldsymbol{Y} \in \mathbb{R}^{n \times 1}$ includes node labels from a class set $\mathcal{C}$.

\begin{figure*}[!t]
    \centering
    \includegraphics[width=1\linewidth]{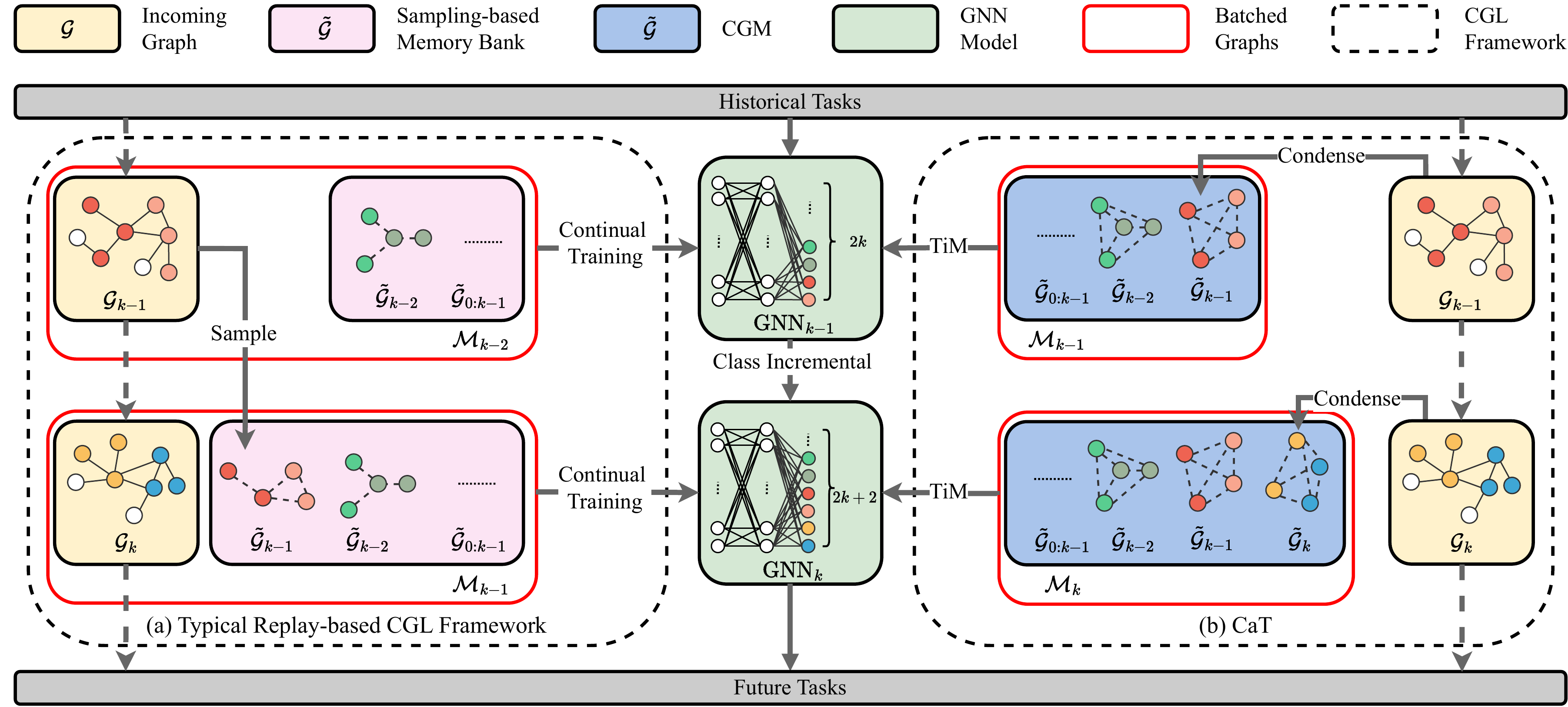}
    \caption{Typical replay-based CGL and CaT in class-IL. (a) A typical replay-based CGL framework trains the model $\text{GNN}_{k-1}$ directly using the memory bank $\mathcal{M}_{k-2}$ and the incoming graph $\mathcal{G}_{k-1}$, and the sampling $\mathcal{\tilde{G}}_{k-1}$ is added to $\mathcal{M}_{k-1}$. (b) CaT condenses the incoming graph $\mathcal{G}_{k-1}$ first to have an updated memory bank $\mathcal{M}_{k-1}$ and trains the model $\text{GNN}_{k-1}$ with $\mathcal{M}_{k-1}$.}
    \label{fig:cat}
\end{figure*}

\subsection{Graph Neural Networks}

Graph Neural Networks (GNNs) are tools for representation learning in node classification problems. The node representation in GNN is calculated by aggregating messages from neighbouring nodes. GNN can be represented as a function:
\begin{equation} \label{eq:gnn}
    \boldsymbol{E} = \text{GNN}_\theta(\boldsymbol{A}, \boldsymbol{X}),
\end{equation}
where $\theta$ is the model parameter and $\boldsymbol{E} \in \mathbb{R}^{n \times d^\prime}$ denotes $d^\prime$-dimensional node embeddings.

\subsection{Graph Condensation}

Graph condensation aims to synthesis a small graph $\mathcal{\tilde{G}}=\{\boldsymbol{\tilde{A}}, \boldsymbol{\tilde{X}}, \boldsymbol{\tilde{Y}}\}$ for a large graph $\mathcal{G}=\{\boldsymbol{A}, \boldsymbol{X}, \boldsymbol{Y}\}$. The model trained with the synthetic graph is expected to have a similar performance as with the original graph. This objective is:
\begin{equation}
\label{eq:gc}
    \min_{\tilde{\mathcal{G}}} \mathcal{L}(\mathcal{G}; \tilde{\theta}),\quad\text{s.t.\ } \tilde{\theta} = \argmin_\theta \mathcal{L}(\tilde{\mathcal{G}}; \theta),
\end{equation}
where $\mathcal{L}$ is a task-related loss function, e.g., cross-entropy, and $\theta$ is the parameter of GNN.

\subsection{Node Classification in CGL}

In node classification of CGL, a model is required to handle $K$ tasks $\{\mathcal{T}_1, \mathcal{T}_2,... \mathcal{T}_{K}\}$. For the $k_{th}$ task $\mathcal{T}_k$, an incoming graph $\mathcal{G}_k$ arrives and the model needs to be updated with $\mathcal{G}_k$ while be tested on all previous graphs and the incoming graph. Following~\cite{twp, ergnn, ssm}, the setting is transductive learning and can be easily extended to inductive learning.

CGL problem has two different continual settings, task incremental learning (task-IL) and class incremental learning (class-IL). In task-IL, the model is only required to distinguish nodes in the same task. While in class-IL, the model is required to classify nodes from all tasks together. Class-IL is more challenging, and this paper focuses on this setting while also reporting the overall performance under task-IL.

\subsection{Imbalanced learning in replay-based CGL methods}

Normally, for the replay-based CGL methods, when storing replayed graphs in the memory bank, there is a budget $b$ limiting the maximum node number for every replayed graph. In the training phase, replayed graphs and the incoming graph are used to train the model together. However, when the size of the incoming graph is significantly larger than the budget, the model will overemphasis on the incoming graph.

Current replay-based CGL methods use weighted loss to tackle the imbalance issue, which combines the current task $\mathcal{T}_k$ loss $\mathcal{L}(\mathcal{G}_{k};\theta_k)$ and the replayed graph loss $\mathcal{L}(\mathcal{M}_{k-1};\theta_k)$. For example, ER-GNN ~\cite{twp} calculates weight by graph size:
\begin{equation}
\label{eq:er-gnn}
\begin{split}
    \ell_\text{ER-GNN} &= \frac{n_{1:k-1}}{n_{k}+n_{1:k-1}} \mathcal{L}(\mathcal{G}_{k};\theta_k) \\
    &+ \frac{n_{k}}{n_{k}+n_{1:k-1}} \mathcal{L}(\mathcal{M}_{k-1};\theta_k),
\end{split}
\end{equation}
where $n_{k}$ and $n_{1:k-1}$ are the node number of the incoming graph $\mathcal{G}_{k}$ and of graphs in memory bank respectively. When $n_{1:k-1} < n_{k}$, the $\mathcal{L}(\mathcal{M}_{k-1};\theta_k)$ are assigned with a larger scale factor. The learning focus will be on the memory bank.

SSM~\cite{ssm} balances the size of each class:
\begin{equation}
\label{eq:ssm}
    \ell_\text{SSM} = \sum_{c \in \mathcal{C}_{k}}\frac{1}{n_{c}} \mathcal{L}_c(\mathcal{G}_{k};\theta_k) + \sum_{c \in \mathcal{C}_{0:k-1}}\frac{1}{n_{c}} \mathcal{L}_c(\mathcal{M}_{k-1};\theta_k),
\end{equation}
where $n_{c}$ is the number of nodes belonging to the class $c$ and the learning will focus on graphs with less nodes. 

Both balancing methods are faced with training problems during the continual learning process. In the class-IL setting, the imbalanced learning problem is mainly caused by the different sample sizes between different classes. ER-GNN will restrict the model from learning on the memory bank once the size of the entire memory bank exceeds the size of the incoming graph significantly. For example, in the Arxiv dataset, SSM will compromise the performance of the current task when the task is sensitive for directly scaling down the training loss. To keep the best performance of SSM, the balancing method using Equation (Eq.)~\ref{eq:ssm} will only be effective on extremely imbalanced datasets (e.g., Reddit and Products).

\section{Methodology}
\label{sec:method}

This section describes the details of the proposed Condense and Train (CaT) framework. The comparisons between CaT and existing replay-based CGL methods are shown in Fig.~\ref{fig:cat}. Existing replay-based CGL methods directly leverage the incoming graph for updating and storing the sampling of the incoming graph in the memory bank. CaT first condenses the incoming graph and updates the model with the condensed graphs instead of the whole incoming graph.

\subsection{Condensed Graph Memory}

Condensed graph memory (CGM) is a memory bank that stores condensed synthetic graphs to approximate the historical data distribution. In this section, we develop graph condensation with distribution matching, aiming to maintain a similar data distribution for the synthetic data as the original data. This approach serves as a replayed graph generation method.

For Task $\mathcal{T}_k$, the incoming graph $\mathcal{G}_k=\{\boldsymbol{A}_k,\boldsymbol{X}_k, \boldsymbol{Y}_k\}$, a condensed graph $\mathcal{\tilde{G}}_k=\{\boldsymbol{\tilde{A}}_k, \boldsymbol{\tilde{X}}_k, \boldsymbol{\tilde{Y}}_k \}$ is generated by graph condensation. Compared with Eq.~\ref{eq:gc}, under the distribution matching scheme, the objective function of graph condensation here can be reformulated as follows:
\begin{equation}
    \mathcal{\tilde{G}}_k^* = \argmin_{\mathcal{\tilde{G}}_k} \text{Dist}(\mathcal{G}_k, \mathcal{\tilde{G}}_k),
\end{equation}
where $\text{Dist}(\cdot,\cdot)$ function calculates the distance between two graphs. Using distribution matching, the distance between two graphs is measured in the embedding space, where both graphs are encoded by the same graph encoder $\text{GNN}_\theta$:
\begin{align}
    \mathcal{\tilde{G}}_k^* &= \argmin_{\mathcal{\tilde{G}}_k} \text{Dist}(\text{GNN}_{\theta_k}(\boldsymbol{A}_k, \boldsymbol{X}_k), \text{GNN}_{\theta_k}(\boldsymbol{\tilde{A}}_k, \boldsymbol{\tilde{X}}_k))\\
    &=\argmin_{\mathcal{\tilde{G}}_k} \text{Dist}(\boldsymbol{E}_k, \boldsymbol{\tilde{E}}_k),
\end{align}
where ${\mathcal{\tilde{G}}}_k^*=\{\boldsymbol{\tilde{A}}_k^*, \boldsymbol{\tilde{X}}_k^*,\boldsymbol{\tilde{Y}}_k^*\}$ is the optimal replayed graph with distribution close to the distribution of the incoming graph. Maximum mean discrepancy (MMD) is used to empirically calculate the distribution distance between two graphs. The objective is to find an optimal $\mathcal{\tilde{G}}_k$ for MMD:
\begin{equation}
\label{eq:loss}
    \ell_{\text{MMD}} = \sum_{c \in \mathcal{C}_k} r_{c} \cdot || \text{Mean}(\boldsymbol{E}_{k,c}) - \text{Mean}(\boldsymbol{\tilde{E}}_{k,c}) ||^2,
\end{equation}
where $\mathcal{C}_k$ is the set of classes of nodes in $\mathcal{G}_k$, $\boldsymbol{E}_{c,k}$ and $\boldsymbol{\tilde{E}}_{c,k}$ are the embedding matrix of the incoming graph and condensed graph, respectively, where all nodes' labels are $c_k$, and $r_{c,k} = \frac{|\boldsymbol{E}_{c,k}|}{|\boldsymbol{E}_k|}$ is the class ratio for class $c_k$. $|\cdot|$ is the number of rows in a matrix. $\text{Mean}(\cdot)$ is the mean vector of the node embeddings.

To efficiently operate the condensation procedure, the random GNN encoders are employed here without training the GNNs. The objective of the distribution matching is to minimise the embedding distance in different embedding spaces given by GNNs with random parameters $\theta_p$:
\begin{equation} \label{eq:gcdm-objective}
    \min_{\mathcal{\tilde{G}}_{k}} \sum_{\theta_p \sim \Theta} \ell_{\text{MMD},\theta_p},
\end{equation}
where $\Theta$ indicates the whole parameter space. The overall procedure of CGM is shown in Algorithm~\ref{alg:gc}. 

With the limit of a budget as $b$, node labels $\boldsymbol{\tilde{Y}} \in \mathcal{C}_k^{b}$ for the condensed graph is initialised and kept as the same class ratio as the original graph (i.e., for any class $c_k$, $r_{k,c} \approx \tilde{r}_{k,c}$). Random sampling from the incoming graph is used to initialise the condensed node features $\boldsymbol{\tilde{X}}_k \in \mathbb{R}^{b \times d}$ at the beginning based on the assigned label. The initialisation can also be implemented as random noise.

\begin{algorithm}[!t]
\SetAlgoLined
\caption{Condensed Graph Memory of $\mathcal{T}_k$}
\label{alg:gc}
\KwIn{Incoming graph $\mathcal{G}_k = \{\boldsymbol{A}_k,\boldsymbol{X}_k, \boldsymbol{Y}_k\}$, budget $b_k$ for the replayed graph $\mathcal{\tilde{G}}_k$}
\KwOut{$\mathcal{\tilde{G}}_k = \{\boldsymbol{\tilde{A}}_k,\boldsymbol{\tilde{X}}_k, \boldsymbol{\tilde{Y}}_k\}$}
Initialise labels of condensed graph $\boldsymbol{\tilde{Y}}_k$ by sampling from the label distribution of $\mathcal{G}_k$\;
Initialise node features of condensed graph $\boldsymbol{\tilde{X}}_k$ by randomly sampling $b_k$ node features from $\boldsymbol{X}_k$\;
Initialise the adjacency matrix $\boldsymbol{\tilde{A}}_k$ with only self-loops\;
\For{$p \gets 1$ \KwTo $P$}{
    Initialise parameter of graph encoder $\theta_p$\;
    \For{$c_k \gets 0$ \KwTo $|\mathcal{C}_k|-1$}{
        $E_{c,k} = \text{GNN}_{\theta_p}(\boldsymbol{A}_k,\boldsymbol{X}_k)$\ \Comment*[r]{Eq.~\ref{eq:gnn}}
        $\tilde{E}_{c,k} = \text{GNN}_{\theta_p}(\boldsymbol{\tilde{A}}_k,\boldsymbol{\tilde{X}}_k)$\ \Comment*[r]{Eq.~\ref{eq:gnn}}
        Calculate $\ell_{\text{MMD},\theta_p}$ according to Eq.~\ref{eq:loss}\\
        $\boldsymbol{\tilde{X}}_k \gets \boldsymbol{\tilde{X}}_k - \eta \nabla_{\boldsymbol{\tilde{X}}_k} \ell_{\text{MMD},\theta_p}$ \Comment*[r]{$\eta$ is learning rate}
    }
}
\end{algorithm}

\subsection{Train in Memory} 

In continual learning, the vanilla replay-based CGL methods are faced with an imbalanced learning problem. When the size of the incoming graph is significantly larger than that of replayed graphs, the model is hard to balance the learning of knowledge from the historical graphs and the incoming graph. The previous attempts for balance are based on the loss scaling. A general form of Eq.~\ref{eq:er-gnn} and~\ref{eq:ssm} can be represented as:
\begin{equation}
    \label{eq:rb}
    \ell_\text{replay} = \alpha\mathcal{L}(\mathcal{G}_{k};\theta_k) + \beta\mathcal{L}(\mathcal{M}_{k-1};\theta_k),
\end{equation}
where most effort is dedicated to $\alpha$ and $\beta$ according to the imbalance scale, which inevitably compromises the performance.

In CaT, since CGM has the ability to condense a graph without compromising the performance, it is reasonable to tackle the imbalance problem by using the condensed incoming graph instead of the whole incoming graph. To incorporate this beneficial characteristic of condensed graphs into the continual learning for balanced training, when the incoming graph $\mathcal{G}_k$ arrived, the condensed graph $\mathcal{\tilde{G}}_k$ is firstly generated, which is then used to update the previous memory $\mathcal{M}_{k-1}$:
\begin{equation}
\label{eq:mem-update}
    \mathcal{M}_{k}=\mathcal{M}_{k-1} \cup \mathcal{\tilde{G}}_k.
\end{equation}

Instead of training with $\mathcal{M}_{k-1}$ and $\mathcal{G}_k$ to deal with the imbalanced issue, CaT will update the model based on $\mathcal{M}_{k}$:
\begin{equation}
\label{eq:cat}
\begin{split}
    \ell_\text{CaT} &= \mathcal{L}(\mathcal{M}_{k};\theta_k)\\
    &=\mathcal{L}(\mathcal{\tilde{G}}_{k};\theta_k) + \mathcal{L}(\mathcal{M}_{k-1};\theta_k).
\end{split}
\end{equation}

This process is named Train in Memory (TiM) since the model only trains with replayed graphs in the memory bank.

In summary, the proposed CaT framework uses graph condensation to generate small and effective replayed graphs and applies the TiM scheme to solve the imbalanced learning in CGL. The overall procedure of CaT is shown in Algorithm~\ref{alg:cat}.

\begin{algorithm}[!t]
\SetAlgoLined
\caption{Overall procedure of CaT}
\label{alg:cat}
\KwIn{A streaming of tasks $\{\mathcal{T}_1, \mathcal{T}_2, ..., \mathcal{T}_K\}$}
\KwOut{$\text{GNN}_{K}$}
Initialise a CGL model $\text{GNN}_0$\;
Initialise an empty memory bank $\mathcal{M}_0$\;
\For{$k \gets 1$ \KwTo $K$}{
    Extract incoming graph $\mathcal{G}_k$ from $\mathcal{T}_k$\;
    Obtain $\mathcal{\tilde{G}}_k$ by CGM\Comment*[r]{Algorithm~\ref{alg:gc}} 
    $\mathcal{M}_k=\mathcal{M}_{k-1} \cup \mathcal{\tilde{G}}_k$ \Comment*[r]{Eq.~\ref{eq:mem-update}}
    Update $\text{GNN}_{k-1}$ to $\text{GNN}_{k}$ \Comment*[r]{Eq.~\ref{eq:cat}}
}
\end{algorithm}

\section{Experiments}
\label{sec:exp}

\subsection{Setup}

\begin{table}[!t]\centering
\caption{Dataset statistics.}\label{tab:dataset}
\begin{tabular}{lrrrrrr}\toprule
Dataset &Nodes &Edges &Features &Classes &Tasks \\\midrule
CoraFull &19,793 &130,622 &8,710 &70 &35 \\
Arxiv &169,343 &1,166,243 &128 &40 &20 \\
Reddit &227,853 &114,615,892 &602 &40 &20 \\
Products &2,449,028 &61,859,036 &100 &46 &23 \\
\bottomrule
\end{tabular}
\end{table}

\paragraph{Datasets} Following the previous work~\cite{hpn, ssm,cglb}, four datasets for node classification tasks are used in experiments, CoraFull~\cite{corafull}, Arxiv~\cite{ogb}, Reddit~\cite{sage} and Products~\cite{ogb}. CoraFull and Arxiv are both citation networks. Reddit is a post-to-post graph. The Products dataset is the co-purchase network. Table~\ref{tab:dataset} shows the statistics of these datasets.

Each dataset is split into a series of tasks focusing on the node classification problem. Each task includes nodes of two unique classes as an incoming graph. In each task, 60\% nodes are chosen as training nodes, 20\% nodes are for validation, and 20\% are for testing. Class-IL is the main focus of the experiment since it is more challenging than task-IL, although overall performance in the task-IL setting will also be reported. In the continual update phase, the model can only access the newly incoming graph and the memory bank. In the testing phase, the model is required to be evaluated with test graphs from all previous tasks. There are no inter-task edges between any two tasks. In the task-IL setting, the output dimension of the model is set to two at all times. In the class-IL setting, since the total class number is not given, the output dimension is incremental as new tasks are coming.

\paragraph{Baselines}
The following baselines are compared:
\begin{itemize}
    \item \textbf{Finetuning} is the lower bound baseline by updating the model only with newly incoming graphs.
    \item \textbf{Joint} is the ideal upper bound situation where the memory bank contains all historical incoming graphs.
    \item\textbf{EWC}~\cite{ewc} applies quadratic penalties to the model weights that are important to the previous tasks.
    \item \textbf{MAS}~\cite{mas} utilises a regularisation term for parameters sensitive to the model performance of historical tasks.
    \item \textbf{GEM}~\cite{gem} modifies the gradients using the informative data stored in memory.
    \item \textbf{TWP}~\cite{twp} preserves the topological information for previous tasks by a regularisation term.
    \item \textbf{LwF}\cite{lwf} distils the knowledge from the old model to the new model to keep the previous knowledge.
    \item \textbf{HPNs}~\cite{hpn} redesign the conventional graph embedding generation for the task-IL setting by maintaining three-level prototypes. Although HPNs are recently published, this baseline is only reported in task-IL experiments.
    \item \textbf{ER-GNN}~\cite{ergnn} samples the informative nodes from incoming graphs into the memory bank.
    \item \textbf{SSM}~\cite{ssm} stores the sparsified incoming graph in the memory bank for future replay.
\end{itemize}

\begin{table*}[!t]\centering
\caption{Overall results for class-IL setting without inter-task edges. All replay-based CGL methods have a budget ratio of 0.01. BWT is also called average forgetting (AF). The bold results are the best performance excluding Joint, and the underlined results are the best baselines excluding Joint. $\uparrow$ denotes the greater value represents greater performance.}\label{tab:classIL}
\begin{tabular}{l|r|rr|rr|rr|rr}\toprule
\multirow{2}{*}{Category} &\multirow{2}{*}{Methods} &\multicolumn{2}{c|}{CoraFull} &\multicolumn{2}{c|}{Arxiv} &\multicolumn{2}{c|}{Reddit} &\multicolumn{2}{c}{Products} \\\cmidrule{3-10}
& &AP (\%) $\uparrow$ &BWT (\%) $\uparrow$ &AP (\%) $\uparrow$ &BWT (\%) $\uparrow$ &AP (\%) $\uparrow$ &BWT (\%) $\uparrow$ &AP (\%) $\uparrow$ &BWT (\%) $\uparrow$ \\\midrule
Lower bound&Finetuning &2.2±0.0 &-96.6±0.1 &5.0±0.0 &-96.7±0.1 &5.0±0.0 &-99.6±0.0 &4.3±0.0 &-97.2±0.1 \\\midrule
\multirow{4}{*}{Regularisation} &EWC &2.9±0.2 &-96.1±0.3 &5.0±0.0 &-96.8±0.1 &5.3±0.6 &-99.2±0.7 &7.6±1.1 &-91.7±1.4 \\
&MAS &2.2±0.0 &-94.1±0.6 &4.9±0.0 &-95.0±0.7 &10.7±1.4 &-92.7±1.5 &10.1±0.6 &-89.0±0.5 \\
&GEM &2.5±0.1 &-96.6±0.1 &5.0±0.0 &-96.8±0.1 &5.3±0.5 &-99.3±0.5 &4.3±0.1 &-96.8±0.1 \\
&TWP &\underline{21.2±3.2} &\underline{-67.4±1.6} &4.3±1.1 &-93.0±8.3 &9.5±2.0 &-35.5±5.5 &6.8±3.5 &-64.3±12.8 \\\midrule
Distillation &LWF &2.2±0.0 &-96.6±0.1 &5.0±0.0 &-96.8±0.1 &5.0±0.0 &-99.5±0.0 &4.3±0.0 &-96.8±0.2 \\\midrule
\multirow{2}{*}{Replay} &ER-GNN &4.0±0.7 &-94.3±0.9 &30.8±0.6 &-68.3±0.7 &31.8±4.0 &-71.2±4.2 &39.5±1.3 &-48.2±1.4 \\
&SSM &16.2±2.8 &-82.1±2.9 &\underline{35.1±1.8} &\underline{-63.7±1.9} &\underline{51.6±6.4} &\underline{-50.3±6.7} &\underline{62.7±0.5} &\underline{-22.1±0.5} \\
\midrule
Full dataset &Joint &85.3±0.1 &-2.7±0.0 &63.5±0.3 &-15.7±0.4 &98.2±0.0 &-0.5±0.0 &72.2±0.4 &-5.3±0.5 \\
\midrule
\midrule
Ours &CaT &\textbf{64.5±1.4} &\textbf{-3.3±2.6} &\textbf{66.0±1.1} &\textbf{-13.1±1.0} &\textbf{97.6±0.1} &\textbf{-0.2±0.2} &\textbf{71.0±0.2} &\textbf{-4.8±0.4} \\
\bottomrule
\end{tabular}
\end{table*}

\begin{table*}[!t]\centering
\caption{Overall results in task-IL setting without inter-task edges. $^*$The results of HPNs are from the original paper, and only Arxiv and Products are provided here.}\label{tab:taskIL}
\begin{tabular}{l|r|rr|rr|rr|rr}\toprule
\multirow{2}{*}{Category} &\multirow{2}{*}{Methods} &\multicolumn{2}{c|}{CoraFull} &\multicolumn{2}{c|}{Arxiv} &\multicolumn{2}{c|}{Reddit} &\multicolumn{2}{c}{Products} \\\cmidrule{3-10}
& &AP (\%) $\uparrow$ &BWT (\%) $\uparrow$ &AP (\%) $\uparrow$ &BWT (\%) $\uparrow$ &AP (\%) $\uparrow$ &BWT (\%) $\uparrow$ &AP (\%) $\uparrow$ &BWT (\%) $\uparrow$ \\\midrule
Lower bound&Finetuning &51.0±3.4 &-46.2±3.5 &67.1±5.2 &-31.3±5.6 &57.1±7.4 &-44.6±7.8 &56.4±3.8 &-42.4±4.0 \\\midrule
\multirow{4}{*}{Regularisation} &EWC &87.4±2.2 &-9.1±2.2 &85.6±7.7 &-11.9±8.1 &85.5±3.3 &-14.8±3.5 &90.3±1.8 &-6.8±1.9 \\
&MAS &93.0±0.3 &\underline{-0.7±0.5} &83.8±6.9 &-12.0±7.8 &99.0±0.1 &\underline{0.0±0.0} &\underline{\textbf{95.9±0.1}} &0.0±0.0 \\
&GEM &\underline{94.3±0.6} &-2.1±0.5 &\underline{94.7±0.1} &-2.3±0.2 &\underline{99.3±0.1} &-0.3±0.1 &86.9±0.9 &-10.6±0.9 \\
&TWP &87.9±1.9 &-4.9±0.6 &77.1±7.3 &-3.5±5.4 &74.1±5.5 &-1.5±0.5 &75.5±4.4 &-4.9±6.4 \\\midrule
Distillation &LwF &64.7±1.1 &-32.3±1.2 &60.2±5.8 &-38.6±6.2 &62.4±3.5 &-39.1±3.7 &50.1±0.7 &-49.3±0.8 \\\midrule
Architecture &$\text{HPNs}^*$ &- &- &85.8±0.7 &\underline{\textbf{0.6±0.9}} &- &- &80.1±0.8 &\underline{\textbf{2.9±1.0}}\\\midrule
\multirow{2}{*}{Replay} &ER-GNN &54.2±1.0 &-43.1±1.1 &92.2±0.3 &-4.9±0.3 &94.3±0.5 &-5.6±0.5 &83.5±0.4 &-14.3±0.5 \\
&SSM &78.7±1.1 &-17.9±1.2 &93.3±0.4 &-3.6±0.4 &99.2±0.2 &-0.5±0.2 &94.6±0.5 &-2.7±0.4 \\\midrule
Full dataset &Joint &97.2±0.0 &0.2±0.1 &96.7±0.0 &-0.1±0.1 &99.7±0.0 &0.0±0.0 &95.7±0.7 &-0.2±0.7 \\
\midrule
\midrule
Ours &CGM &\textbf{95.3±0.3} &\textbf{-0.3±0.3} &\textbf{95.8±0.2} &0.2±0.1 &\textbf{99.4±0.0} &\textbf{0.1±0.1} &95.6±0.3 &0.1±0.3 \\
\bottomrule
\end{tabular}
\end{table*}

\paragraph{Evaluation Metrics} When the model is updated after Task $\mathcal{T}_k$, all previous tasks from $\mathcal{T}_{1}$ to $\mathcal{T}_{k}$ are evaluated. A lower triangular performance matrix $\boldsymbol{M} \in \mathbb{R}^{K \times K}$ is maintained, where $m_{i,j}$ denotes the classification accuracy of Task $\mathcal{T}_j$ after learning from Task $\mathcal{T}_i$ ($i \leq j$). Additionally, the following metrics are used to compare different methods comprehensively.

\textbf{Average performance (AP)} measures the average model performance after learning from Task $\mathcal{T}_k$:
\begin{equation}
    \text{AP}_k = \frac{1}{k}\sum^k_{i=1}m_{k,i}.
\end{equation}

\textbf{Mean of average performance ($\overline{\text{AP}}$)}~\cite{classIL-survey} denotes the average performance of model snapshots in the continual leanring process:
\begin{equation}
    \overline{\text{AP}} = \frac{1}{k}\sum^k_{i=1}\text{AP}_i.
\end{equation}

\textbf{Backward transfer (BWT)}~\cite{cl-survey1} (also known as the average forgetting (AF)) indicates how the training process of the current task affects the previous tasks. The larger number implies that training the current task will have a greater impact on historical tasks. A negative or a positive number implies a negative or a positive impact, respectively:
\begin{equation}
    \text{BWT}_k = \frac{1}{k-1}\sum_{i=1}^{k-1} (m_{k,i} - m_{i,i}).
\end{equation}

\paragraph{Implementation} The budget ratio represents the proportion of the memory bank to the total number of nodes in the entire training set, and the budget for every task is evenly assigned. By default, the budget ratio for the Joint baseline is 1 because it stores every training data in its memory. For example, 0.01 is the budget ratio in most following experiments, and the size of the memory bank becomes 1\% of the size of the entire training data. Although the budget is set to a real number instead of a ratio of the entire training set in more piratical scenarios, the budget ratio is used in the experiments for keeping fairness and comparing the efficiency of different memory banks. Unless otherwise specified, for the replay-based method, the default budget ratio is 0.01.

GCN is the default backbone model. For CGM, a 2-layer GCN with a 512-dimensional hidden layer is used to encode all four datasets, and other graph encoders are evaluated as well. The learning rate for the condensed feature matrix is 0.0001 for CoraFull and 0.01 for other datasets. For the node classification problem, a 2-layer GCN with a 256-dimensional hidden layer and a class number-dependent output layer is used as the node classifier in all datasets. Unless otherwise specified, all results are obtained by running three times and reported with the average value and the standard error. All experiments are conducted on one NVIDIA RTX 2080 Ti GPU.

\subsection{Overall Results}
\label{sec:overall-results}
The CaT is compared with all baselines in both class-IL and task-IL settings. AP is used to evaluate the average model performance of all learned tasks at the end of the task streaming, and BWT (also known as average forgetting (AF)) implies the forgetting problem of the model during continual learning. Table~\ref{tab:classIL} shows the overall performance of all baselines and the CaT in the class-IL CGL setting. CaT achieves the state-of-the-art performance compared with all other CGL baselines and can match the ideal Joint performance in the Arxiv, Reddit and Products by only maintaining a synthetic memory bank whose budget ratio is only 0.01. Besides, the results show that CaT has a smaller BWT, which means CaT cannot only preserve the historical knowledge of the model but reduce the negative effects on the previous tasks while training the current task to alleviate the catastrophic forgetting problem. Although CaT outperforms other baselines in CoraFull, CaT does not reach the Joint performance with two potential reasons: (1) the 0.01 budget ratio for CoraFull limits the replayed graph to four nodes, which is extremely small to contain sufficient information; (2) CoraFull has 35 tasks, which is more than other datasets and difficult to retain historical knowledge.

Other baselines can hardly match the performance of CaT. Finetuning is easy to forget the previous knowledge since it only uses the newly incoming graph to update the model. Regularisation-based methods (e.g., EWC, MAS, GEM, TWP) also have unsatisfactory performance since adding overhead restrictions to the model will lead to bad model plasticity during the long streaming tasks. As a distillation method, LwF hardly handles the class-IL setting in the CGL. ER-GNN does not have reasonable results in all benchmarks for the sampling-based replay methods since there is a severe imbalanced training problem. SSM stores sparsified subgraphs in the memory bank, which can preserve the topological information for the historical graph data. Although SSM has a good performance, it still has a gap to Joint or CaT.

It is worth noting that even though all historical data can be used for training, Joint also has a negative BWT. The reason is that the class-IL setting requires the model to increase the output layer dimension as the new classes emerge, where the model cannot perfectly remember all previous knowledge. The results of HPNs for the class-IL are not provided since it is not designed for the class-IL setting.

Table~\ref{tab:taskIL} shows the overall performance under the task-IL setting. Compared to the class-IL setting, task-IL is much easier, and all baseline methods get reasonable results. The CaT achieves the state-of-the-art performance and can match the Joint method with only a 0.01 budget ratio.

\subsection{Ablation Study}

\begin{table}[!t]\centering
\caption{Ablation study of the CaT framework.}\label{tab:ablation}
\begin{tabular}{r|r|r|r|r|r}\toprule
\multirow{2}{*}{CGM} &\multirow{2}{*}{TiM} &\multicolumn{2}{c|}{CoraFull} &\multicolumn{2}{c}{Arxiv} \\\cmidrule{3-6}
& &AP (\%) $\uparrow$ &BWT (\%) $\uparrow$ &AP (\%) $\uparrow$ &BWT (\%) $\uparrow$ \\\midrule
\xmark &\xmark &16.2±2.8 &-82.1±2.9 &35.1±1.8 &-63.7±1.9 \\
\cmark &\xmark &54.8±1.3 &-41.7±1.1 &34.2±7.0 &-64.3±7.5 \\
\xmark &\cmark &16.8±5.0 &-29.0±6.1 &53.5±1.4 &-16.2±1.4 \\
\cmark &\cmark &64.5±1.4 &-3.3±2.6 &66.0±1.1 &-13.1±1.0 \\\midrule\midrule
\multirow{2}{*}{CGM} &\multirow{2}{*}{TiM} &\multicolumn{2}{c|}{Reddit} &\multicolumn{2}{c}{Products} \\\cmidrule{3-6}
& &AP (\%) $\uparrow$ &BWT (\%) $\uparrow$ &AP (\%) $\uparrow$ &BWT (\%) $\uparrow$ \\\midrule
\xmark &\xmark &51.6±6.4 &-50.3±6.7 &62.7±0.5 &-22.1±0.5 \\
\cmark &\xmark &60.2±3.7 &-41.3±3.9 &71.3±0.2 &-13.5±0.6 \\
\xmark &\cmark &92.1±1.2 &-4.3±1.6 &63.2±0.2 &-9.7±0.6 \\
\cmark &\cmark &97.6±0.1 &-0.2±0.2 &71.0±0.2 &-4.8±0.4 \\
\bottomrule
\end{tabular}
\end{table}

The CaT framework has two key components, CGM and TiM. To study their effectiveness, different CaT variants are evaluated, and the AP and BWT of these variants are reported in Table~\ref{tab:ablation}. The variant without CGM indicates using Random Choice for the memory bank, and the variant without TiM indicates the typical replay-based scheme using the whole incoming graph for training. According to Table~\ref{tab:ablation}, compared to the variant without both components, the variant using CGM improves both AP and BWT in the CoraFull and Arxiv datasets but dropped in the Reddit and Products datasets. The main reason is that CoraFull and Arxiv are small datasets where CGM can easily capture the data distribution. CGM hardly benefits the model for large datasets, especially in the Products dataset, where the scale imbalance problem is dominant. Compared to the variant without TiM, the variant with TiM significantly improves the overall performance, especially on large datasets (e.g., Reddit and Products). This also reflects the effectiveness of TiM for imbalanced graphs.

\subsection{Effectiveness and Efficiency of CGM}

\begin{figure}[!t]
\begin{tabular}{cc}
  \includegraphics[width=42mm]{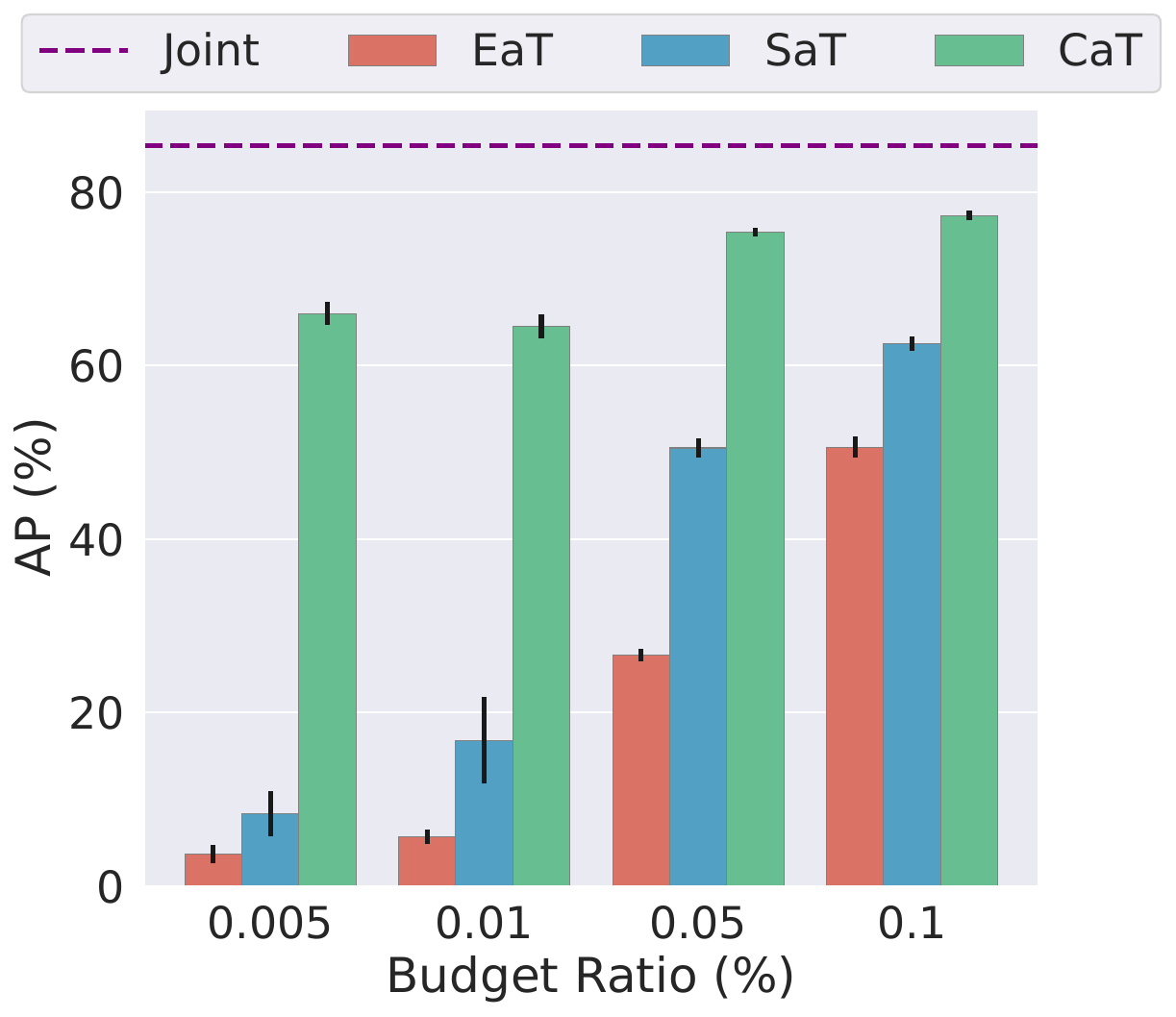} &   \includegraphics[width=42mm]{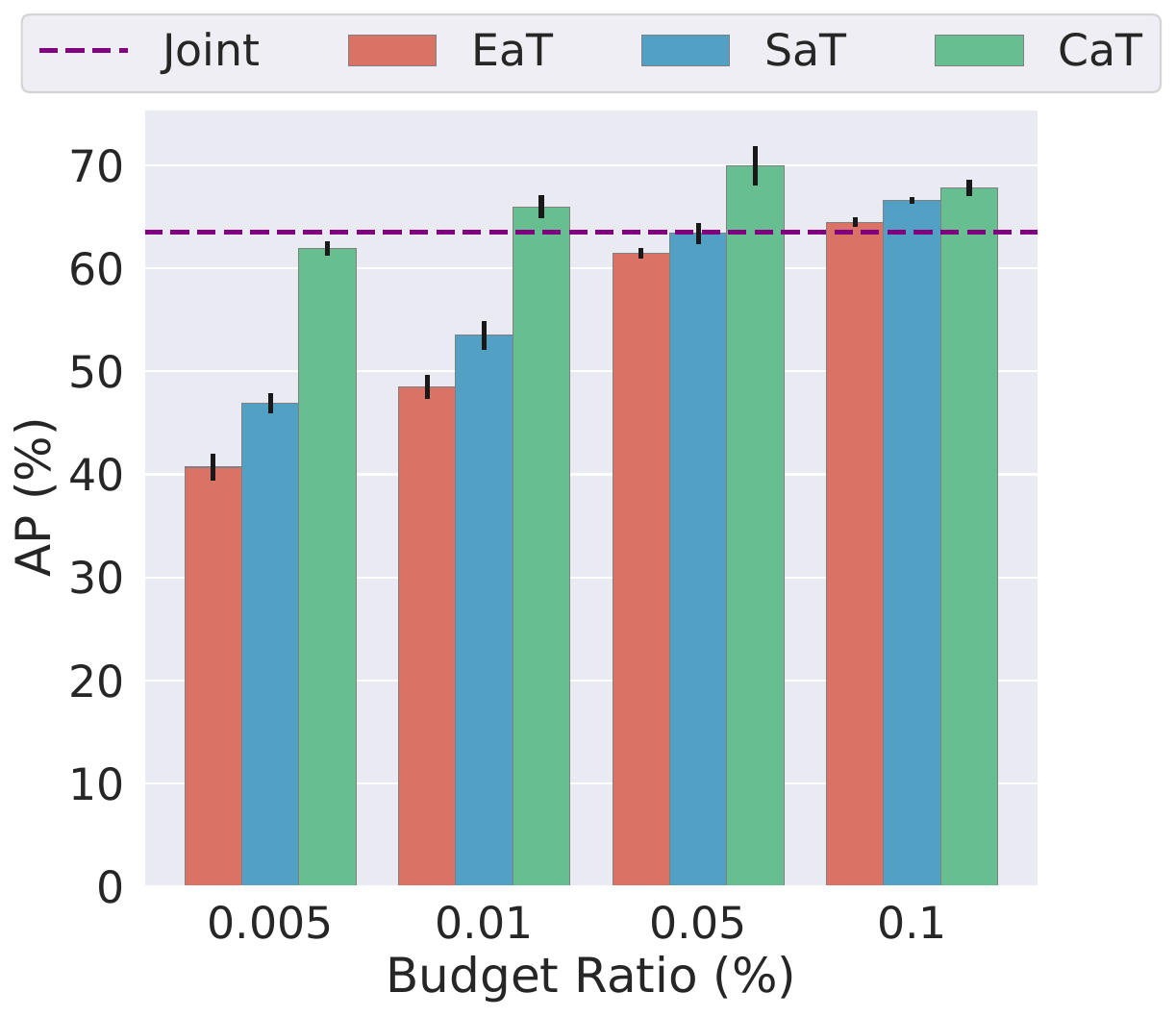} \\
(a) CoraFull & (b) Arxiv \\
\includegraphics[width=42mm]{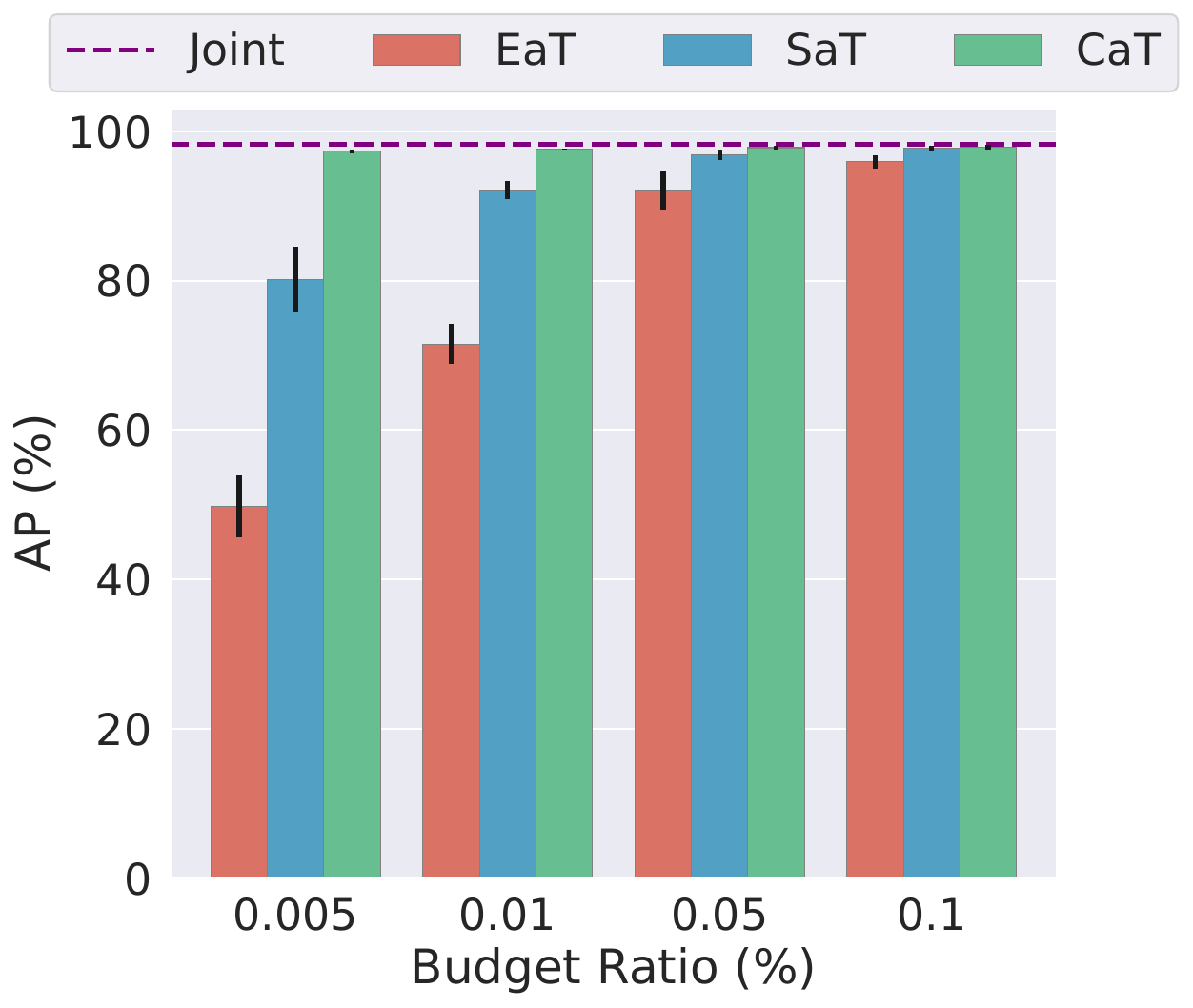} &   \includegraphics[width=42mm]{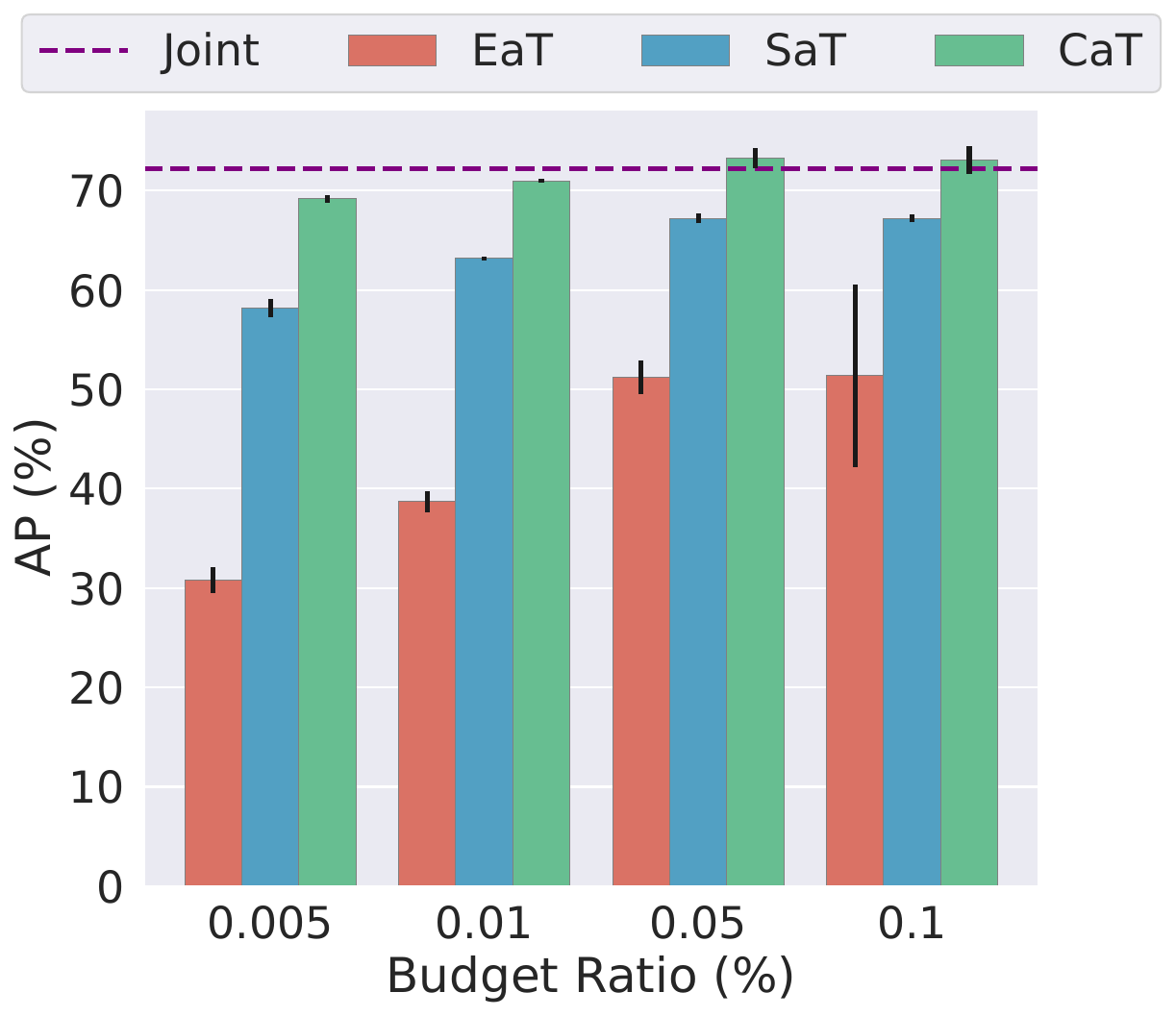} \\
(c) Reddit & (d) Products \\
\end{tabular}
\caption{AP of methods with different budget ratios. All methods use the TiM to avoid imbalanced training for fairness. CGM is more effective and efficient than ER-GNN and SSM.}
\label{fig:budgets}
\end{figure}

To analyse the effectiveness and efficiency of CGM, different memory banks are evaluated with four budget ratios, i.e., 0.005, 0.01, 0.05, and 0.1. Specifically, 0.005 is an extremely limited budget ratio, under which CGM only contains 2-node replayed graphs on the CoraFull dataset. While 0.1 is a large budget ratio, which represents the size of the memory bank equals to 10\% of the size of the entire training set. For a fair comparison with CaT, the TiM scheme is applied for ER-GNN and SSM. EaT and SaT are used to denote the ER-GNN with TiM and SSM with TiM for short. The AP is used here.

\paragraph{Different Memory Banks} Fig.~\ref{fig:budgets} demonstrates that the CGM is more effective than existing sampling-based memory banks. CGM converges to optimal performance much quicker. CGM almost gets the best performance in all evaluated cases. CGM significantly outperforms other sampling-based memory banks when the budget ratio is relatively small (e.g., 0.005, 0.01). Although in the Arxiv dataset, SSM can outperform the CGM when the budget ratio is as large as 0.1, which is impractical for the memory bank. Besides, CGM has a small standard error, demonstrating that CGM is more robust for continual training. In the CoraFull dataset, the AP of 0.01 budget is slightly lower than 0.005. The size of replayed graphs with 0.01 budget (4-node replayed graph) and 0.005 budget (2-node replayed graph) is similar. The only difference is that the 2-node replayed graph has a uniform class distribution, but the 4-node replayed graph keeps the same class distribution as the original graph. The potential reason is the class imbalance problem raised in the small replayed graphs. We leave the balancing between classes as a future research problem. 

\paragraph{Budget Efficiency} The advantage of the condensed graph is to keep the information of the original graph while reducing the graph size significantly. Fig.~\ref{fig:budgets} shows that the CGM method outperforms the sampling-based methods in achieving higher performance within a more limited budget. In all datasets, the sampling-based methods have a huge performance gap to CGM. 0.005-budget ratio CGM can outperform or match the 0.1-budget ratio sampling-based methods.

On the one hand, CGM uses less memory space to accurately approximate the historical data distribution. On the other hand, in the training phase, the model needs to propagate messages in the memory bank. Therefore, a small memory bank can improve both storage and computation efficiency.

\begin{figure}[!t]
\begin{tabular}{cc}
\includegraphics[width=42mm]{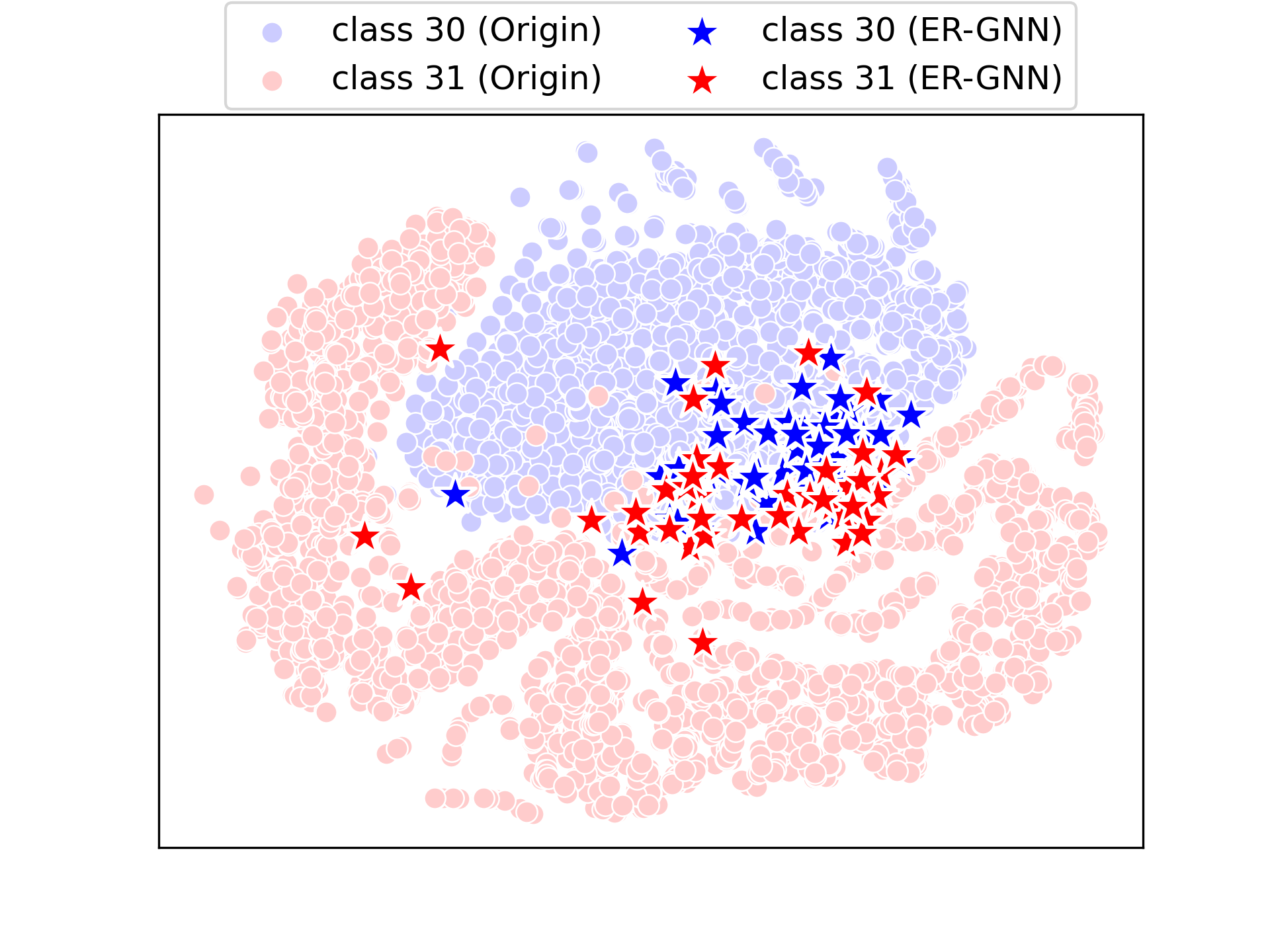} &   
\includegraphics[width=42mm]{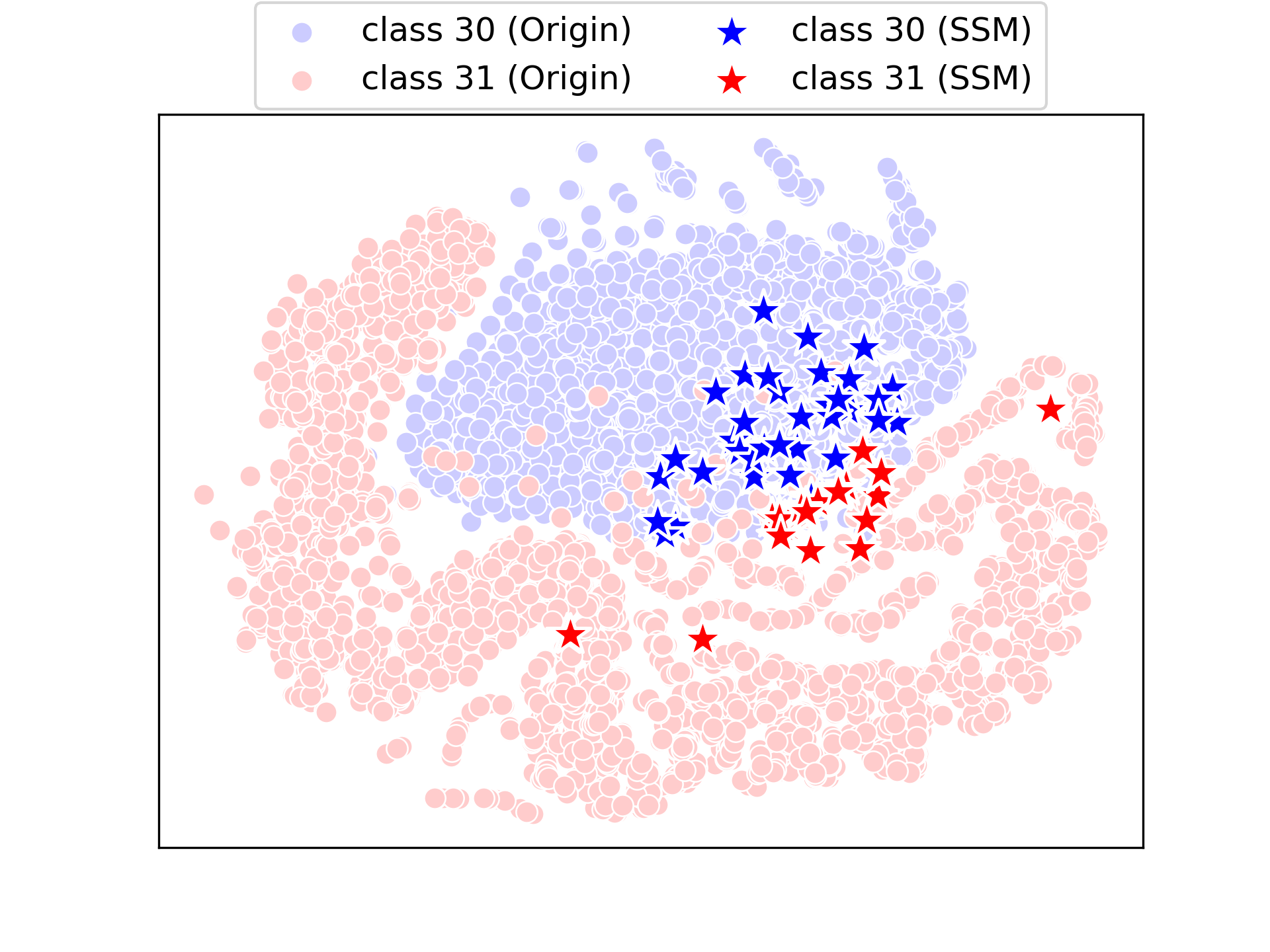} \\
(a) ER-GNN &(b) SSM \\
\includegraphics[width=42mm]{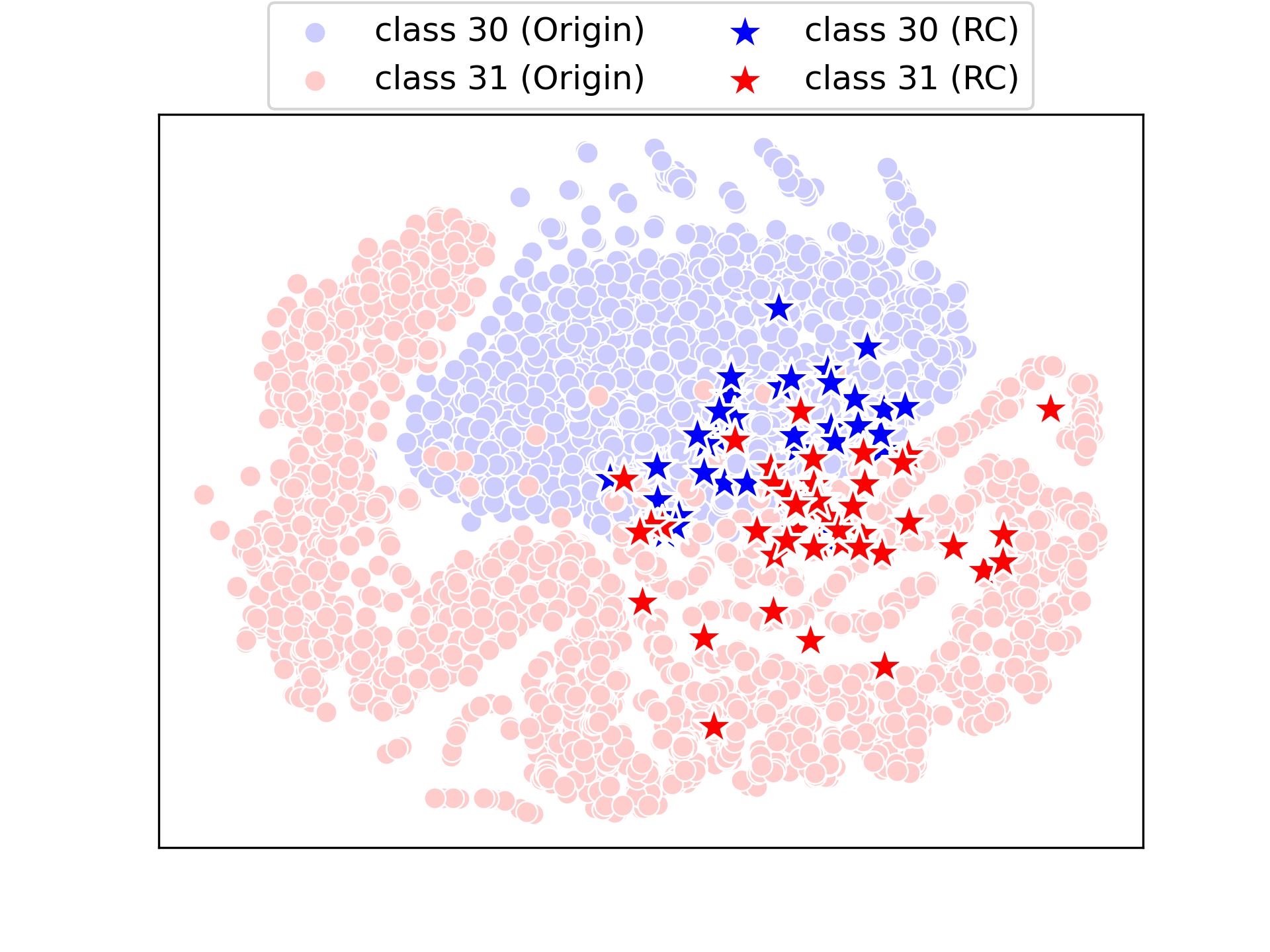} &
\includegraphics[width=42mm]{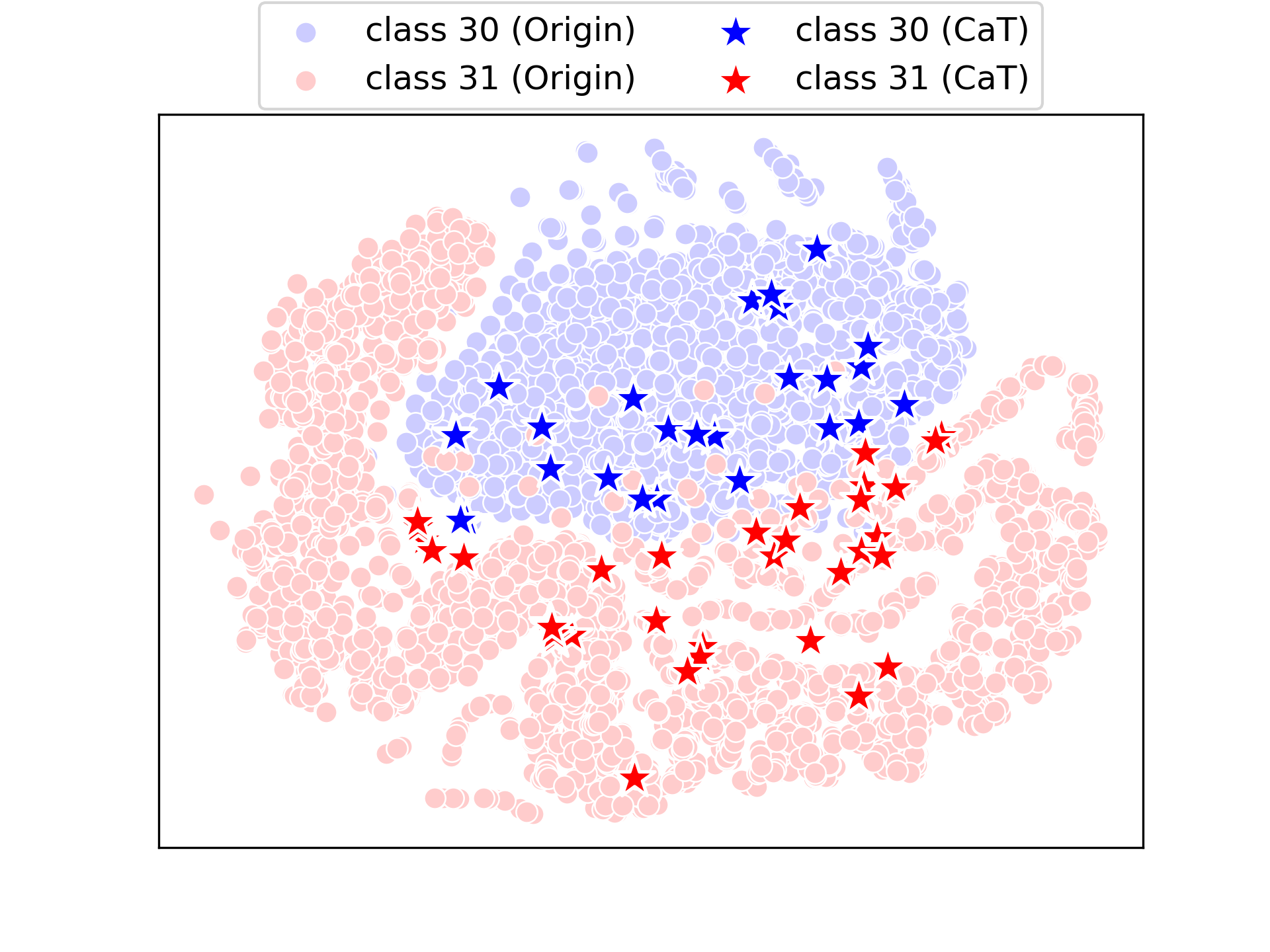} \\
(c) Random Choice &(d) CGM \\
\end{tabular}
\caption{The visualisation of node embeddings from the original graph and the replayed graph of different methods (ER-GNN, SSM, Random Choice and CGM) in Task 15 of Reddit.}
\label{fig:tsne-emb}

\end{figure}
\paragraph{Visualisation} To further explore the effectiveness of CGM. The t-SNE~\cite{tsne} visualises the node embedding of different memory banks in Fig.~\ref{fig:tsne-emb}, which shows the embedding distribution of ER-GNN, SSM, CGM initialisation (Random Choice) and CGM. To keep the same embedding space for reasonable comparisons, all node embeddings are generated with the same graph encoder. The node embeddings in sampling-based memory banks are close to each other, which cannot effectively approximate the complete graph distribution. On the contrary, the embedding of CGM is more diverse and can cover the whole distribution, enabling the classification model to learn a more accurate decision boundary.

\subsection{Balanced Learning with TiM}

\begin{table}[!t]\centering
\caption{$\overline{\text{AP}}$ (\%) of different methods using TiM scheme.}\label{tab:tims}
\begin{tabular}{r|r|r|r|r|r}\toprule
&TiM &CoraFull &Arxiv &Reddit &Products \\\midrule
\multirow{2}{*}{ER-GNN} &\xmark &11.5±0.1 &36.2±0.6 &48.5±1.4 &35.1±0.2 \\
&\cmark &15.7±0.5 &57.1±0.6 &80.0±1.8 &42.8±0.5 \\\midrule
\multirow{2}{*}{SSM} &\xmark &12.0±0.1 &39.4±0.9 &71.0±0.9 &57.5±0.2 \\
&\cmark &36.7±2.5 &62.4±1.2 &92.8±1.3 &74.6±0.3 \\\midrule
\multirow{2}{*}{CGM} &\xmark &12.0±0.1 &39.4±0.9 &71.0±0.9 &65.6±0.1 \\
&\cmark &36.7±2.5 &62.4±1.2 &92.8±1.3 &81.7±0.1 \\
\bottomrule
\end{tabular}
\end{table}

\paragraph{Different Methods with TiM} TiM is a plug-and-play training scheme for all existing replay-based CGL methods. Table ~\ref{tab:tims} shows the mean of AP for different replay-based CGL methods with and without TiM. It clearly shows that the TiM scheme can improve the overall average performance with all memory bank generation methods. The reason is that the TiM can ensure training graphs for the CGL models have a similar size to deal with the imbalanced issue, which can solve the catastrophic forgetting problem. 

\begin{figure*}[!t]
\begin{tabu}{ccccccc}
\includegraphics[width=25mm]{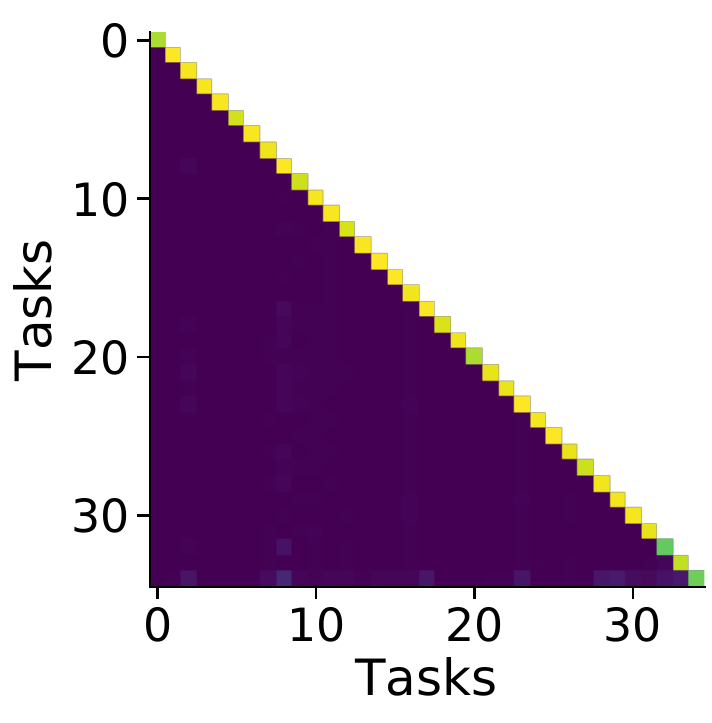} &   \includegraphics[width=25mm]{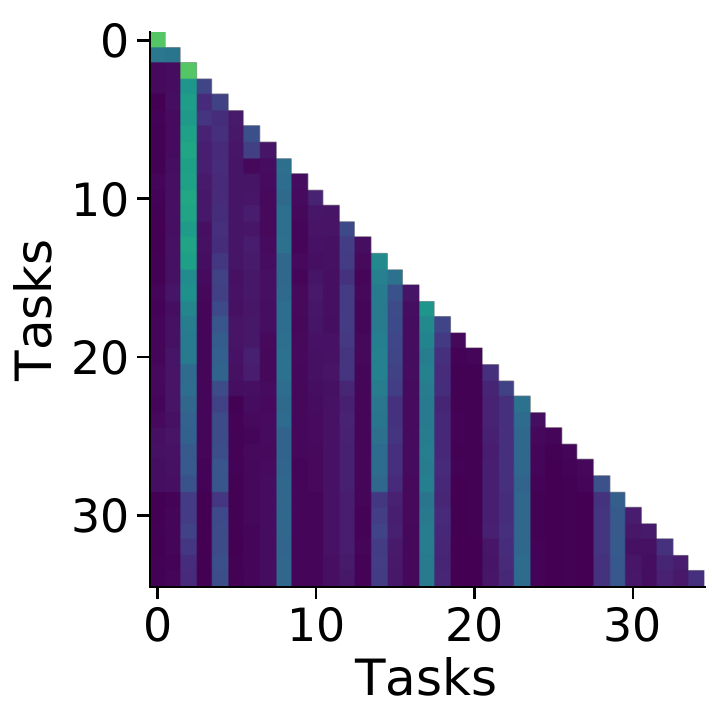} &
\includegraphics[width=25mm]{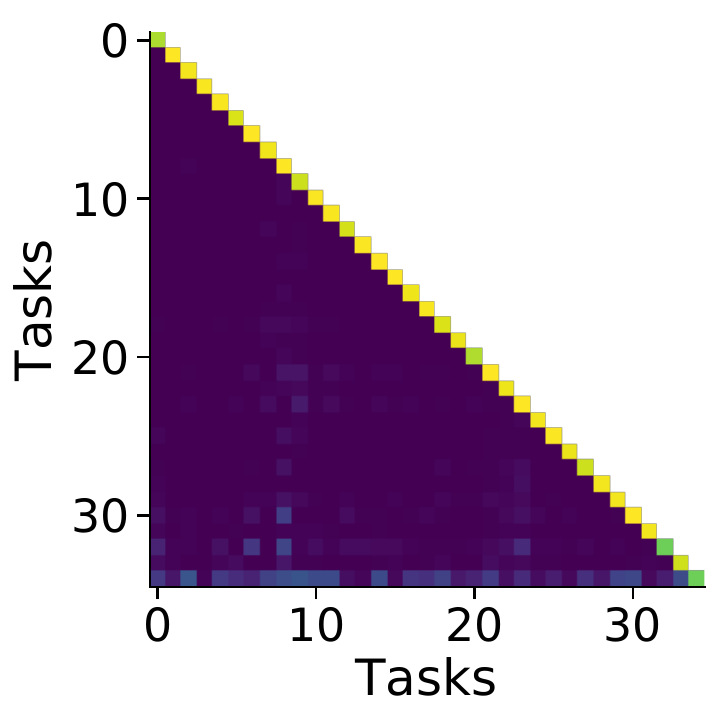} &   \includegraphics[width=25mm]{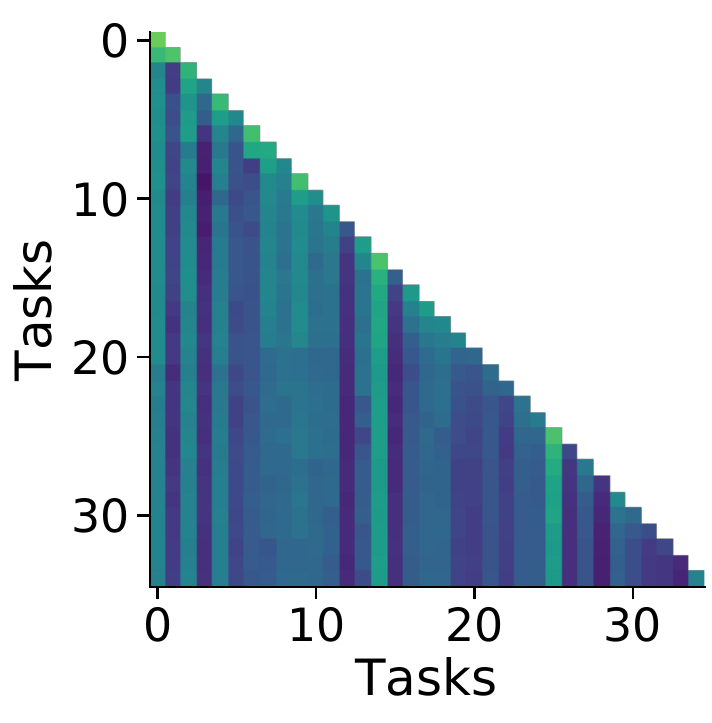} &
\includegraphics[width=25mm]{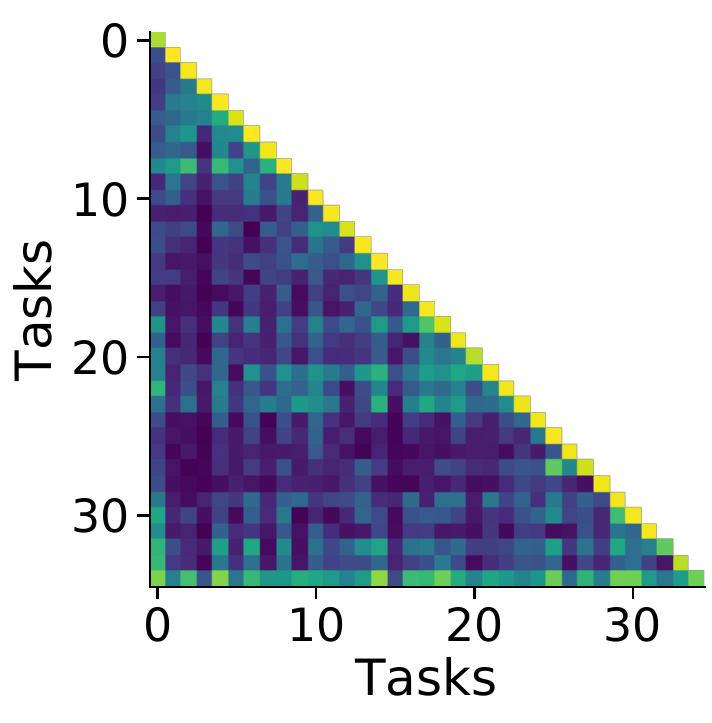} &
\includegraphics[width=25mm]{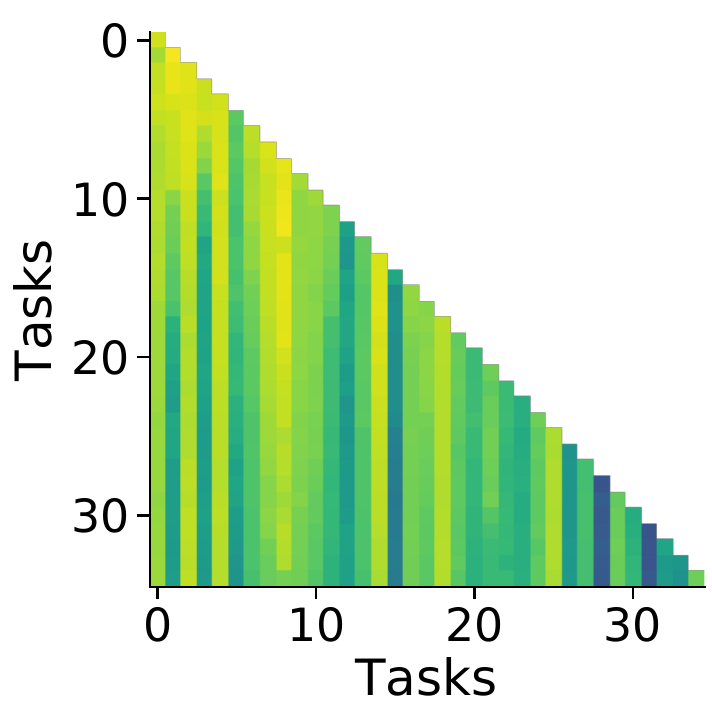} &   \includegraphics[width=5.3mm]{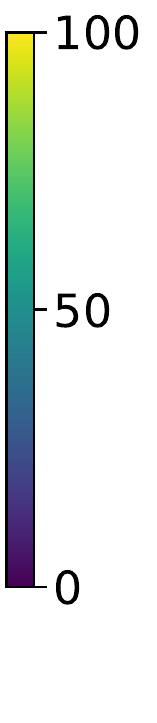} \\
\rowfont{\scriptsize}
ER-GNN & EaT &SSM &SaT &CGM &CaT & \\
\multicolumn{7}{c}{(a) CoraFull} \\
\includegraphics[width=25mm]{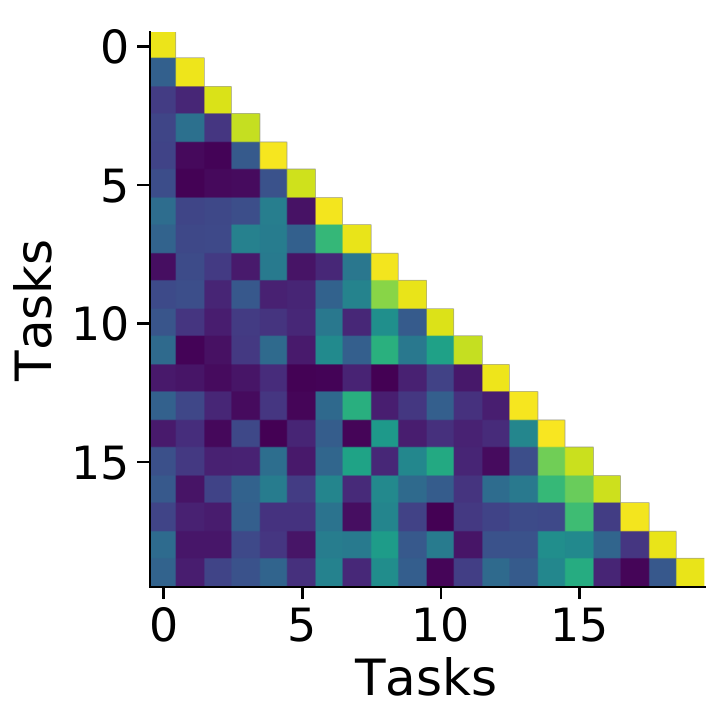} &   \includegraphics[width=25mm]{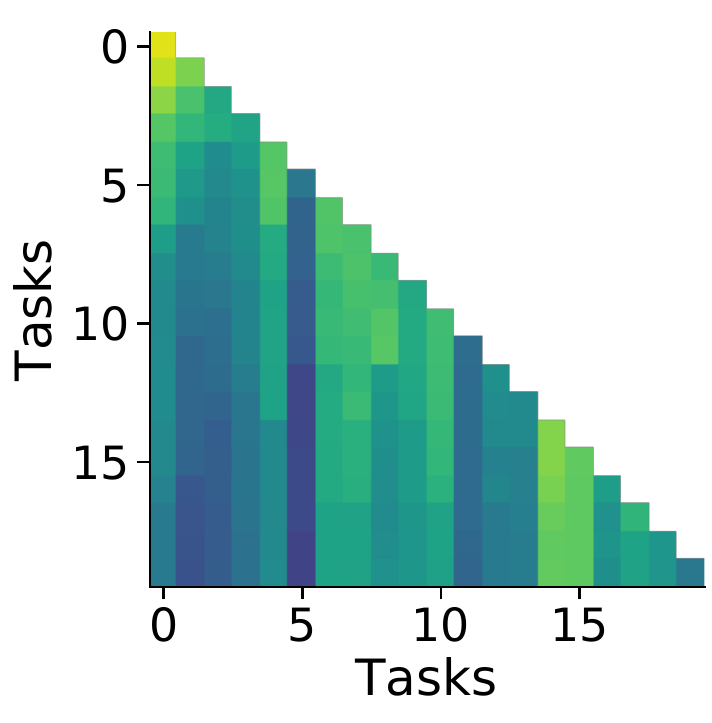} &
\includegraphics[width=25mm]{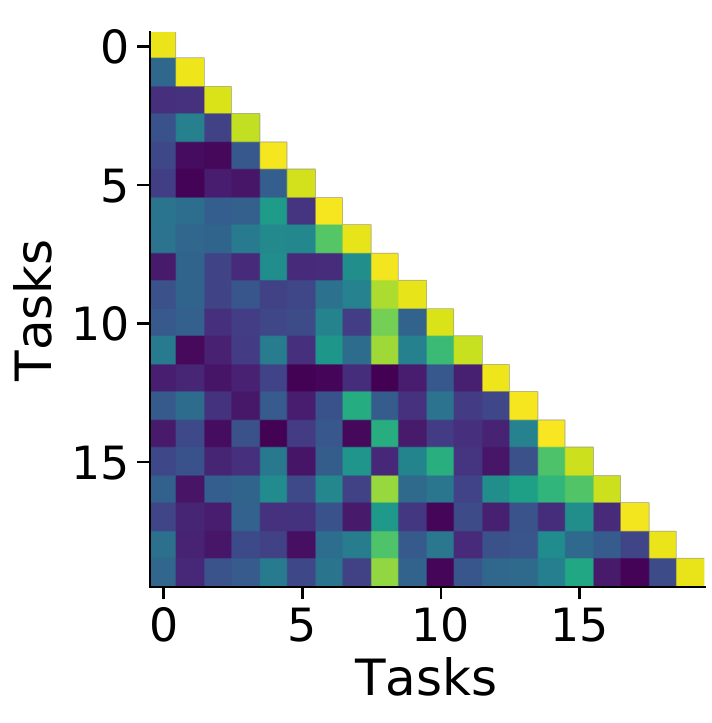} &   \includegraphics[width=25mm]{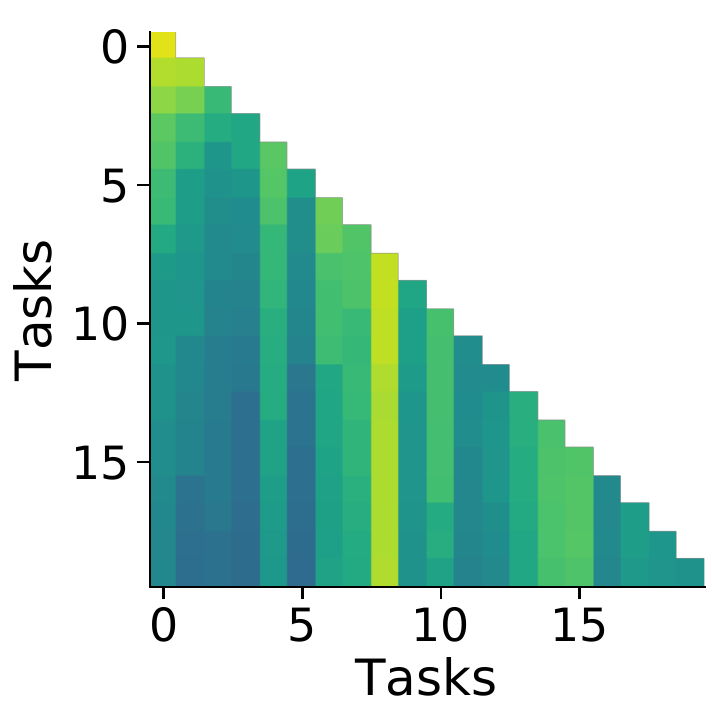} &
\includegraphics[width=25mm]{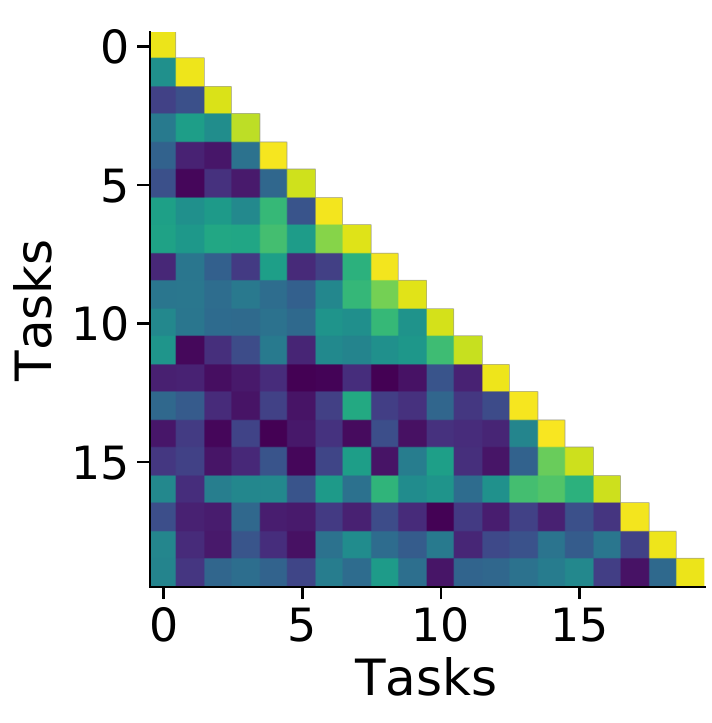} &
\includegraphics[width=25mm]{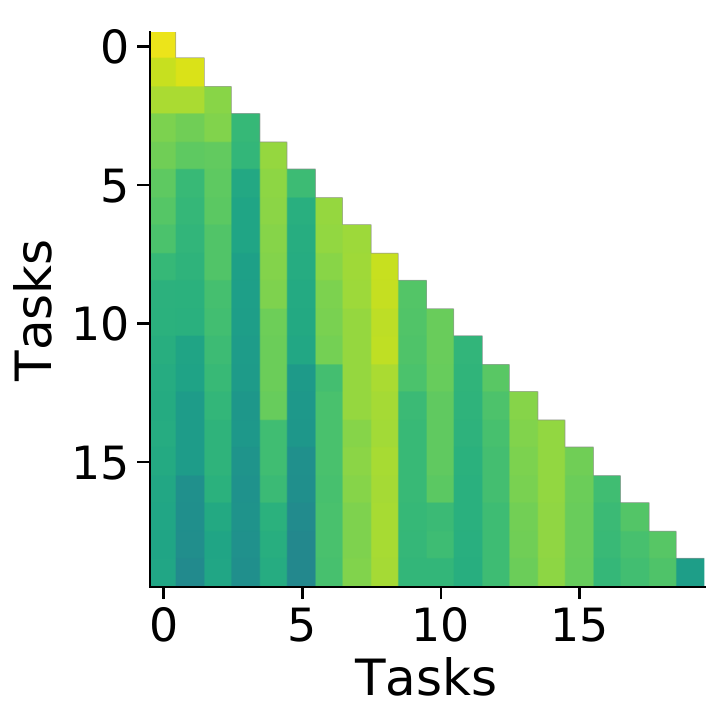} &   \includegraphics[width=5.3mm]{Figures/acc/color_bar.pdf} \\
\rowfont{\scriptsize}
ER-GNN & EaT &SSM &SaT &CGM &CaT & \\
\multicolumn{7}{c}{(b) Arxiv} \\
\includegraphics[width=25mm]{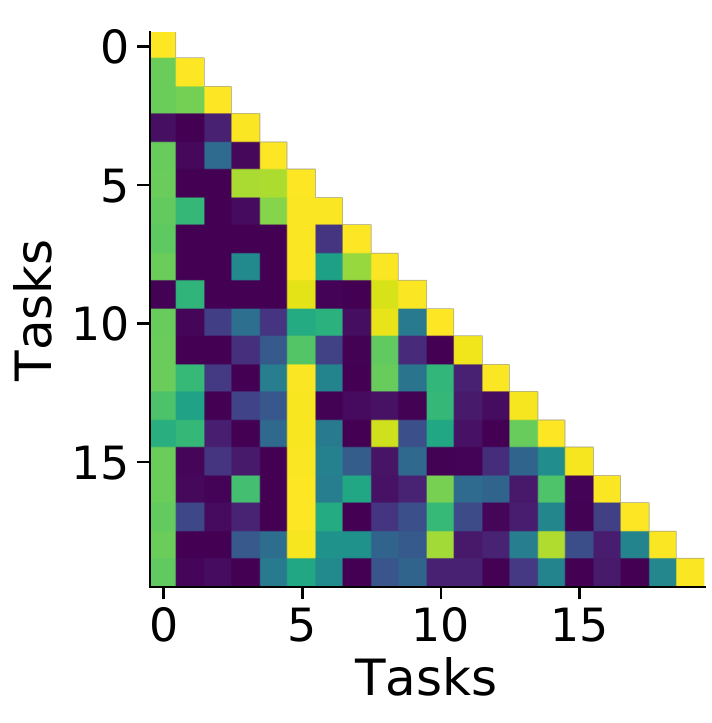} &   \includegraphics[width=25mm]{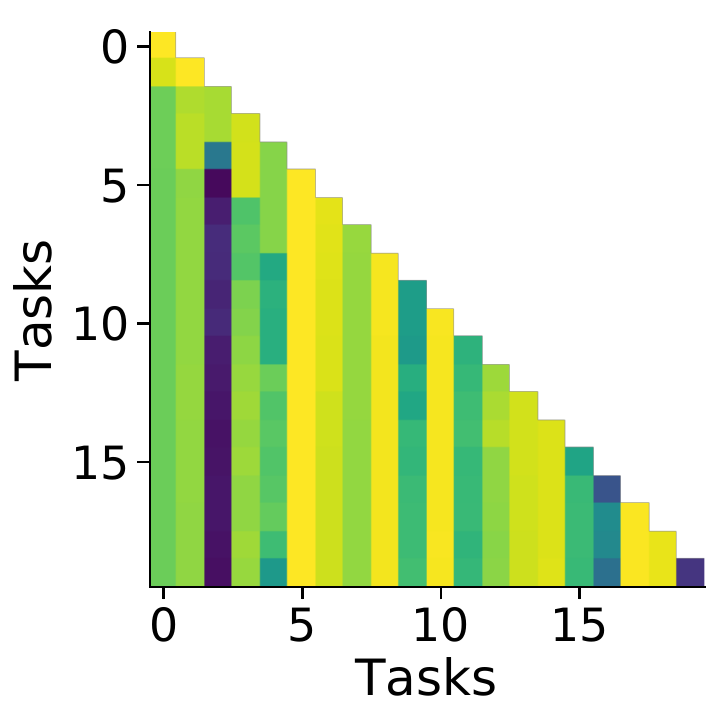} &
\includegraphics[width=25mm]{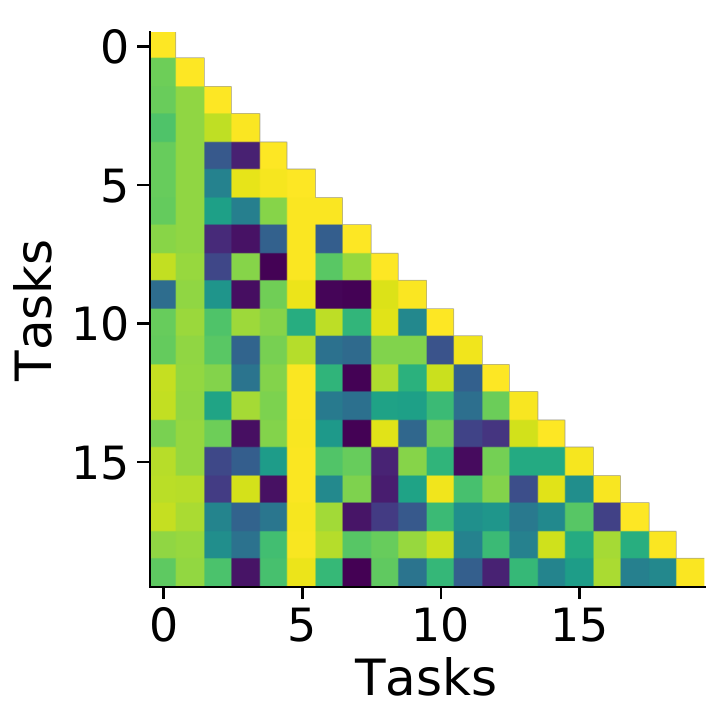} &   \includegraphics[width=25mm]{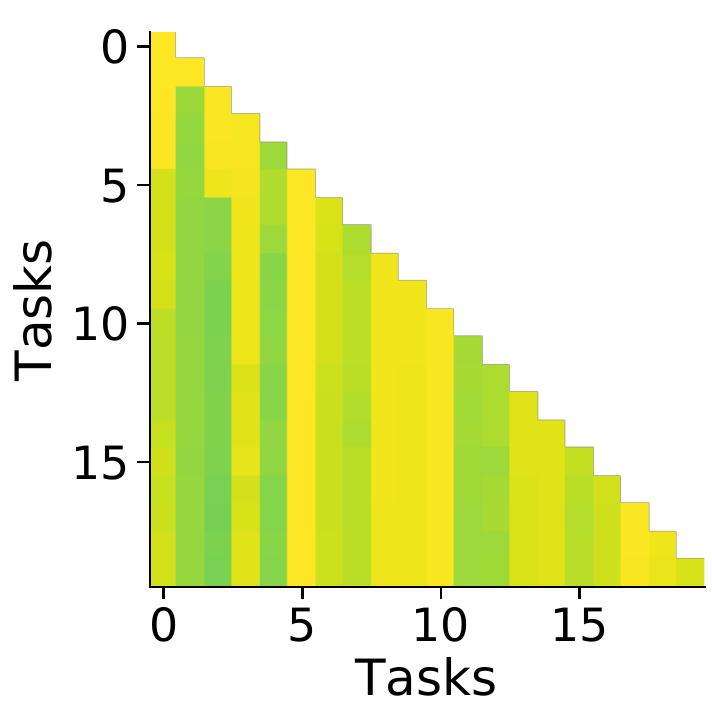} &
\includegraphics[width=25mm]{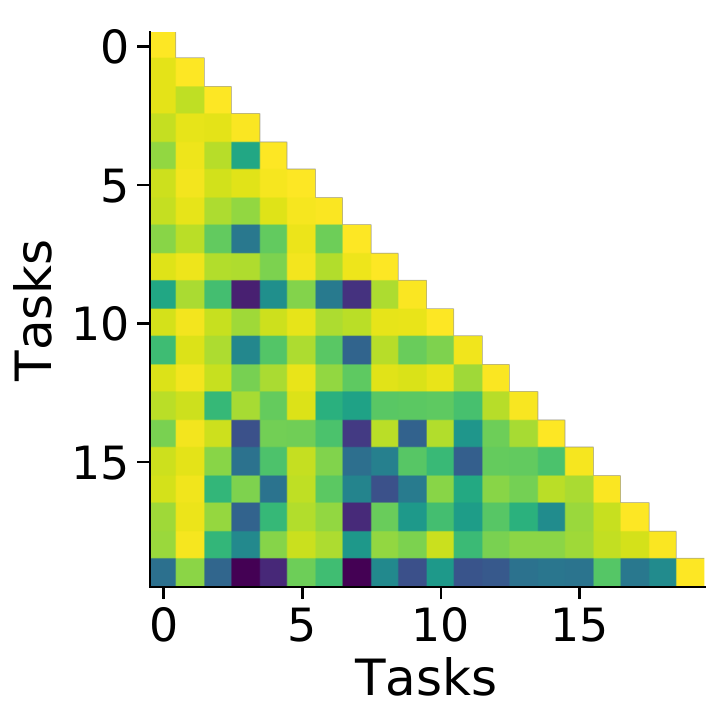} &
\includegraphics[width=25mm]{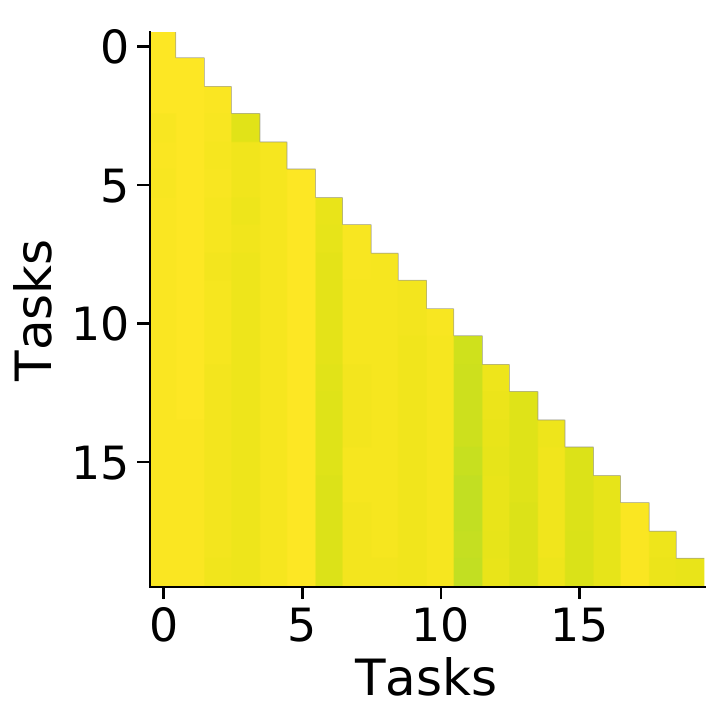} &   \includegraphics[width=5.3mm]{Figures/acc/color_bar.pdf} \\
\rowfont{\scriptsize}
ER-GNN & EaT &SSM &SaT &CGM &CaT & \\
\multicolumn{7}{c}{(c) Reddit} \\
\includegraphics[width=25mm]{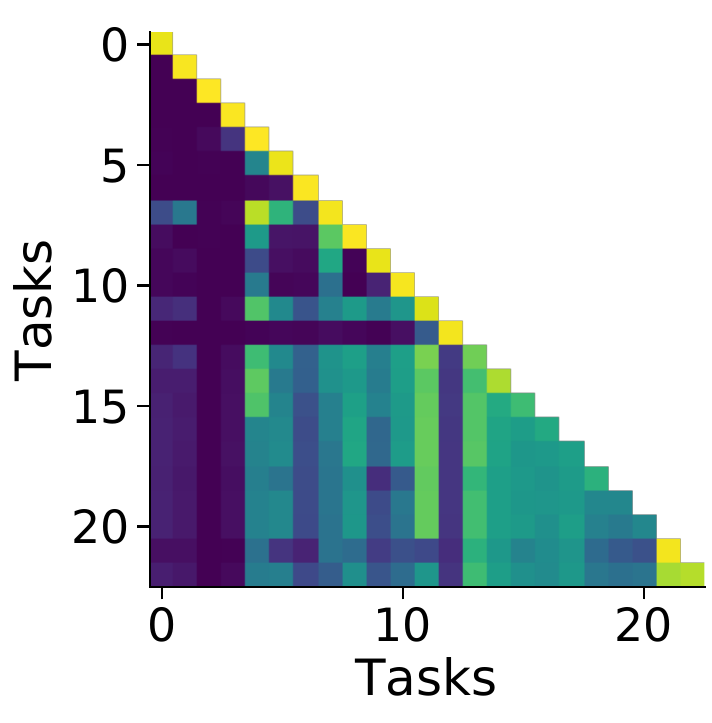} &   \includegraphics[width=25mm]{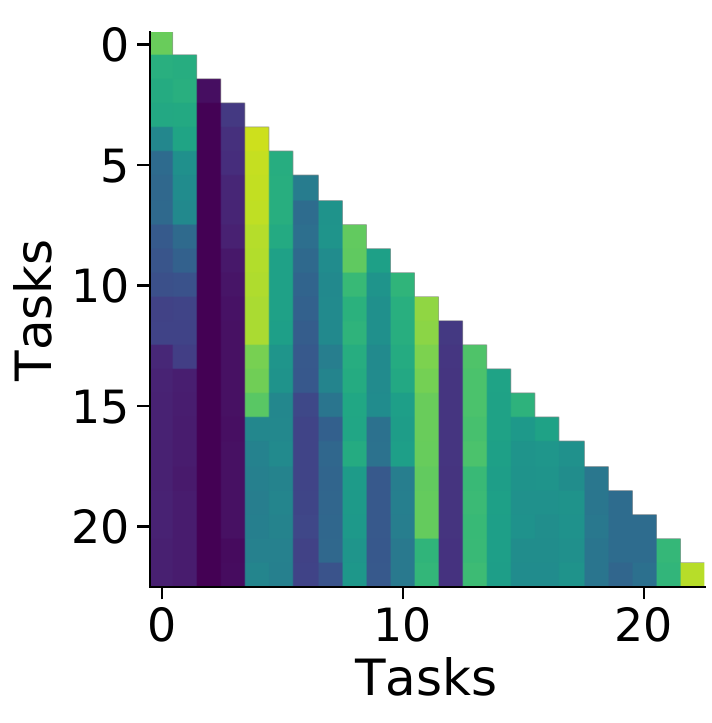} &
\includegraphics[width=25mm]{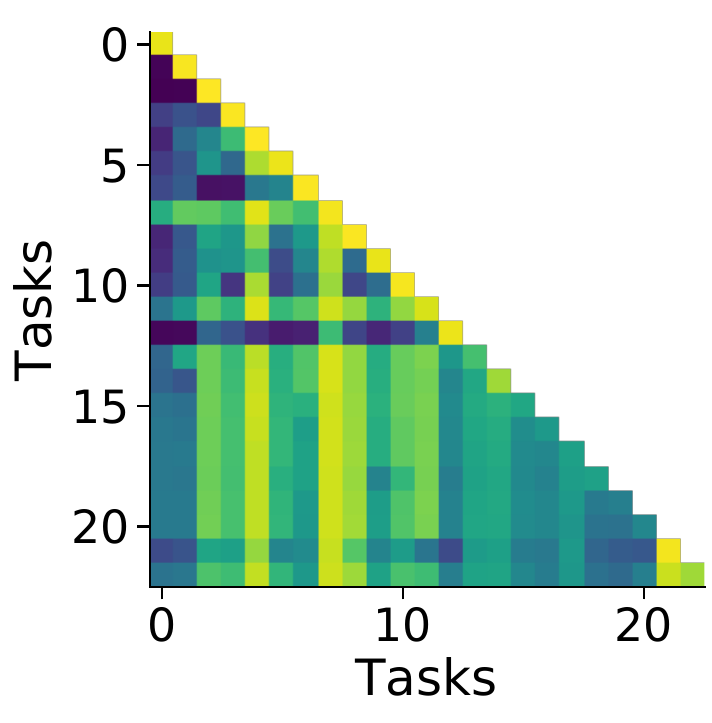} &   \includegraphics[width=25mm]{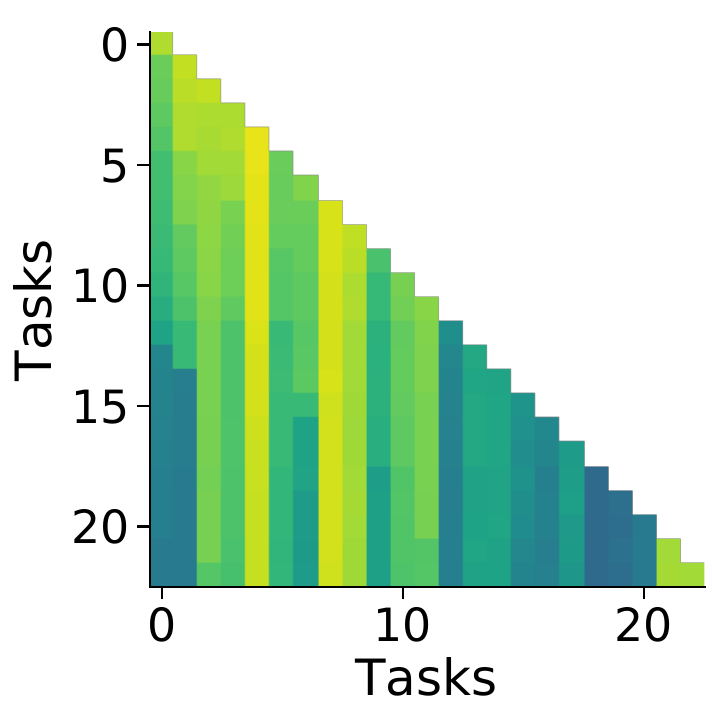} &
\includegraphics[width=25mm]{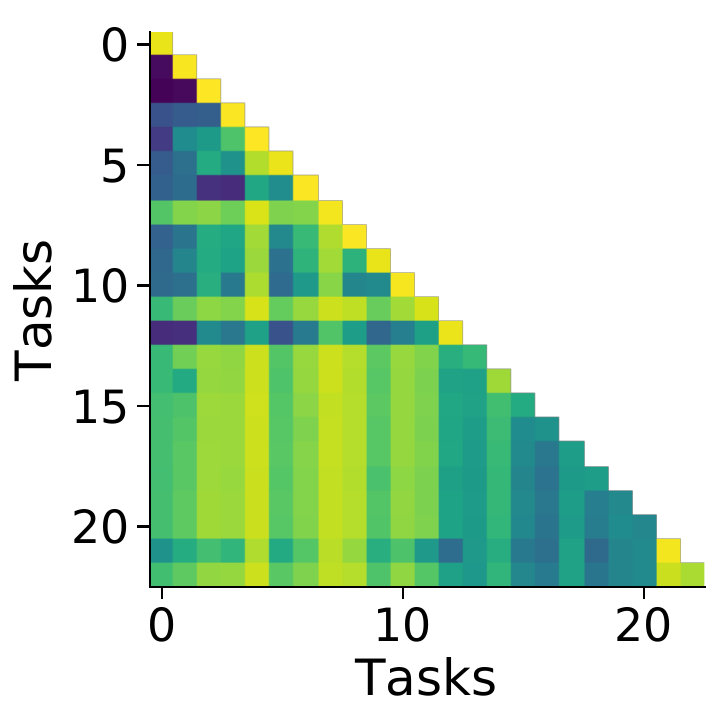} &
\includegraphics[width=25mm]{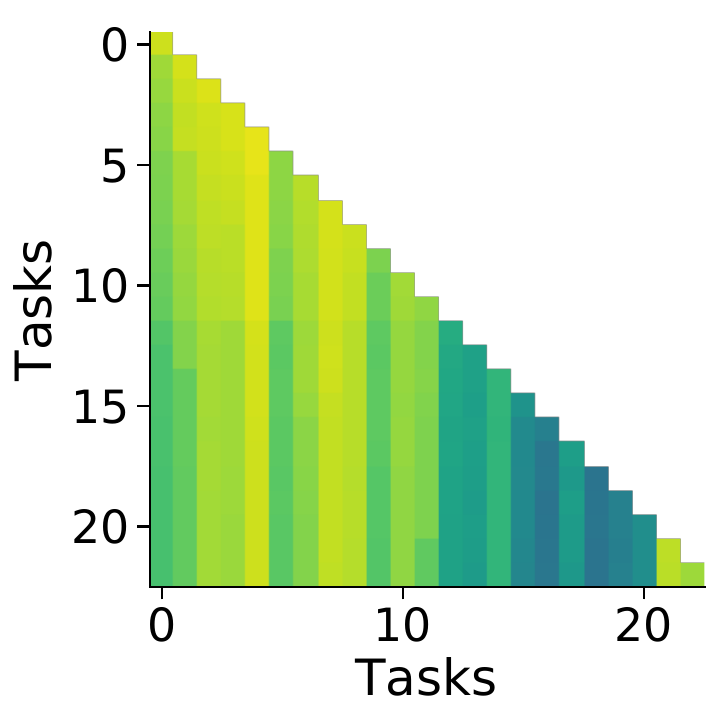} &   \includegraphics[width=5.3mm]{Figures/acc/color_bar.pdf} \\
\rowfont{\scriptsize}
ER-GNN & EaT &SSM &SaT &CGM &CaT & \\
\multicolumn{7}{c}{(d) Products} \\
\end{tabu}
\caption{Performance matrix visualisation of ER-GNN, SSM, CGM and their combination with TiM scheme in CoraFull, Arxiv, Reddit and Products datasets. The coloured square located at the $i_{th}$ row and the $j_{th}$ column denotes the classification accuracy of Task $\mathcal{T}_j$ after model training on Task $\mathcal{T}_i$. Light colour means high accuracy, and dark colour means low accuracy. The $i_{th}$ column from top to bottom can represent the accuracy changes during the model's continual training of Task $\mathcal{T}_i$.}
\label{fig:acc}
\end{figure*}

\paragraph{Visualisation} The performance matrices of ER-GNN, SSM, CGM, and those memory banks training with TiM (i.e., EaT, SaT can CaT) on the CoraFull, Arxiv, Reddit and Products datasets under 0.01 budget ratio are visualised in Fig.~\ref{fig:acc}. All memory banks without TiM struggle with remembering the previous knowledge since the scale gap between the newly incoming graph and replayed graphs in the memory bank. After using the TiM scheme, the performance matrices show the forgetting process slows down (i.e., the colour of each column is not changed a lot), which indicates the catastrophic forgetting problem is alleviated as the imbalanced training issue is tackled.

\begin{table}[!t]\centering
\caption{AP (\%) of different graph encoders.}\label{tab:encoder}
\resizebox{1\linewidth}{!}{
\begin{tabular}{r|rrr|rrr}\toprule
\multirow{2}{*}{} &\multicolumn{3}{c|}{CoraFull} &\multicolumn{3}{c}{Arxiv} \\\cmidrule{2-7}
&0.01 &0.05 &0.1 &0.01 &0.05 &0.1 \\\midrule
SGC &56.7±1.7 &74.9±1.0 &78.5±0.5 &64.2±0.4 &66.6±0.9 &64.3±0.3 \\
GCN &64.5±1.4 &75.4±0.5 &77.3±0.6 &66.0±1.1 &70.0±1.9 &67.8±0.8 \\
\midrule\midrule
\multirow{2}{*}{} &\multicolumn{3}{c|}{Reddit} &\multicolumn{3}{c}{Products} \\\cmidrule{2-7}
&0.01 &0.05 &0.1 &0.01 &0.05 &0.1 \\\midrule
SGC &97.2±0.1 &97.5±0.1 &97.6±0.1 &65.1±0.3 &68.6±0.4 &68.2±0.2 \\
GCN &97.6±0.1 &97.8±0.2 &97.9±0.3 &71.0±0.2 &73.3±1.0 &73.1±1.4 \\
\bottomrule
\end{tabular}}
\end{table}

\subsection{Parameter Sensitivity}
There are several hyperparameters in CGM, including the budget for the replayed graph, which is already evaluated in Fig.~\ref{fig:budgets}. This section will discuss the choice of graph encoders.

\paragraph{Different Graph Encoders} This experiment will compare GCN~\cite{gcn} and SGC~\cite{sgc} as encoders for CGM, while GCN will still be applied in the node classification for CGL. AP is used to measure the effectiveness. All CGM encoders have a 256-dimensional hidden layer and a 128-dimensional output layer. Table~\ref{tab:encoder} shows that for different budget ratios, SGC and GCN can serve as competitive encoders for CGM. One exception is that in CoraFull with the budget ratio of 0.01, the performance of CGM with SGC is much lower than that with GCN. This is possible because 0.01 is a very strict budget ratio for the CoraFull dataset that SGC does not have the sufficient representation ability with such less data. A similar situation happens in the Products dataset, which is a huge and challenging dataset for GNNs. 

\section{Conclusion}
\label{sec:con}
This paper identifies the inefficient sampling-based memory bank and unbalanced continual learning issues in the replay-based CGL methods. To solve these issues, a novel CaT framework is proposed, which includes two key components, CGM and TiM. CGM is a small yet effective memory bank based on the graph condensation. TiM scheme updates the memory bank with the newly incoming graph and continuously trains the model with this memory bank to balance the update. Extensive experiments demonstrate that this framework achieves state-of-the-art performance in task-IL and class-IL settings.

\section*{Acknowledgment}
This work is supported by Australian Research Council CE200100025 and DP230101196.

\bibliographystyle{IEEEtranS}
\bibliography{IEEEabrv,ref}

\end{document}